\documentclass[12pt]{article}
\usepackage{amsmath,amssymb,fullpage,graphicx,amscd,array,amsfonts,mathtools,amsthm}
\usepackage{xcolor}
\usepackage{url}
\usepackage[english]{babel}
\usepackage{graphicx}
\usepackage{longtable}
\usepackage{caption}
\usepackage{appendix}
\usepackage[font=footnotesize]{caption}
\newtheorem{strategy}{Strategy}
\newtheorem{algorithm}{Algorithm}
\newtheorem{step}{Step}[algorithm]
\newtheorem{substep}{}[step]
\newtheorem{substrategy}{}[strategy]
\newtheorem{ruler}{Rule}

\RequirePackage{geometry}
\allowdisplaybreaks

\begin{document}

%
%
%
%
%
\renewcommand{\thefootnote}{\alph{footnote}}

\begin{center}
{\LARGE Automated Generation of Geometric Theorems \\
\vspace{3mm} from Images of Diagrams}\\

\vspace{5mm}

Xiaoyu Chen,\footnote{\it SKLSDE - School of Computer Science and Engineering, Beihang University, Beijing 100191, China. E-mail: franknewchen@gmail.com} Dan Song,\footnote{\it LMIB - School of Mathematics and Systems Science, Beihang University, Beijing 100191, China} and Dongming Wang$^{\rm b,}$\footnote{\it Centre National de la Recherche Scientifique, 3 rue Michel-Ange, 75794 Paris cedex 16, France}
  
\vspace{3mm}

\end{center}

\begin{abstract}
\centering
\vspace{3 true pt}
\begin{minipage}{0.8\textwidth}
\indent
We propose an approach to generate geometric theorems from
electronic images of diagrams automatically. The approach makes use
of techniques of Hough transform to recognize geometric objects and
their labels and of numeric verification to mine basic geometric
relations. Candidate propositions are generated from the retrieved
information by using six strategies and geometric theorems are obtained
from the candidates via algebraic computation. Experiments with a
preliminary implementation illustrate the effectiveness and
efficiency of the proposed approach for generating nontrivial
theorems from images of diagrams. This work demonstrates the feasibility
of automated discovery of profound geometric knowledge from simple
image data and has potential applications in geometric knowledge
management and education.
\end{minipage}
\end{abstract}

\renewcommand{\thefootnote}{\arabic{footnote}}
\setcounter{footnote}{0}

\section{Introduction}\label{intro}

Diagrams have been used to illustrate geometric theorems and problems for thousands of years and can be produced now by using computer programs with ease. A number of software tools developed in the area of dynamic geometry are capable of generating dynamic diagrams automatically from specifications of geometric theorems~\cite{dgs}. This paper tackles the inverse problem: given the electronic image of a diagram, generate the specifications of one or more theorems which the diagram may be used to illustrate. An ideal solution to this problem allows one to generate geometric theorems automatically from their illustrations available in electronic documents and resources.

To be specific, let us restrict our study to plane Euclidean geometry in this paper. The approach we propose to solve the above-stated problem consists of the following main steps.
\begin{enumerate}
\item Retrieve geometric information, mainly geometric objects and their labels, in the given image of diagram using techniques of pattern recognition (in particular Hough transform as discussed in Sections~\ref{objectsrec} and~\ref{labelrec}).
\item Mine geometric relations among the retrieved geometric objects from predetermined potential relations by examining their corresponding algebraic relations using numeric verification (see Section~\ref{relationsrec}).
\item Generate candidate propositions from the mined geometric relations by using six strategies introduced (see Section~\ref{candidateGener}).
\item Obtain theorems from the generated candidate propositions using algebraic methods, first  to rule out false propositions efficiently by checking numeric instances and then to prove the true propositions via symbolic computation (see Sections~\ref{theoremMine} and~\ref{theoremProve}).
\end{enumerate}
These four steps are described in detail in the following two sections. We have implemented the proposed approach. Preliminary experiments with our implementation are reported in Section~\ref{implementation}. Some related work on geometric information retrieval and theorem discovery  is discussed briefly in Section~\ref{relatedwork}. The paper concludes with a few remarks in Section~\ref{conclusion}.

The work presented in this paper demonstrates for the first time the
feasibility of discovering rigorous and profound geometric knowledge
(theorems) from inexact and partial geometric data (images of diagrams)
automatically. This feasibility brings us the hope to build up a
large-scale database of geometric theorems automatically or
semi-automatically by searching diagrams from electronic documents
and resources accessible via the Internet. Theorems collected in
such a database will have standardized formal representations and
are linked to images of diagrams. The processing and management of
theorems in the database, including  searching, organization,
translation (into representations in natural languages or algebraic
expressions), and  degenerate-case handling, would be made easier or
more efficient.

A potential application of our work in education is evident. An extension of the proposed approach to dealing with hand-drawn diagrams could make it possible for students to submit geometric theorems to provers by simply sketching their diagrams on mobile computing devices.

\section{Information Retrieval from Images of Diagrams}
\label{inforetrieval}

Geometric information consists of geometric objects (i.e., shapes
used in geometric diagrams), their labels (i.e., identifiers of the
objects), and geometric relations (i.e., properties and features of
the objects). In what follows, we discuss how to retrieve
information from images of diagrams and how to represent it in a
processable form for theorem generation.

\subsection{Recognizing Basic Geometric Objects}
\label{objectsrec}

In our current investigation we consider the following three types
of basic geometric objects which are used to form most of the
diagrams in plane Euclidean geometry.

\begin{itemize}
\item \textbf{Points.} A point is represented by a pair of coordinates $(x,y)$ in the coordinate system determined by the image of diagram, where the parameters $x$ and $y$ are called \emph{$x$-coordinate} and \emph{$y$-coordinate}, respectively.
\item \textbf{Lines.} A straight line (with no extremes) is represented by \texttt{line}$(P_1,P_2)$, where the parameters $P_1$ and $P_2$ denote two distinct points incident to the line. Similarly, a half line is represented by \texttt{halfline}$(O,P)$,
where the parameter $O$ denotes its initial point and $P$ denotes a
point on it; a segment is represented by
\texttt{segment}$(E_1,E_2)$,  where the parameters $E_1$ and $E_2$
denote the endpoints of the segment. For the sake of convenience, all
straight lines, half lines, and segments, with respective types
\texttt{line}, \texttt{halfline}, and \texttt{segment}, are called
\emph{lines}.
\item \textbf{Circles.} A circle may be represented by \texttt{circle}$(O,r)$, where the parameter $O$ denotes the center of the circle and $r~(>0)$ denotes the radius of the circle, and by \texttt{circle}$(A, B, C)$, where the parameters $A$, $B$, and $C$ denote three distinct points on the circle.
\end{itemize}
A geometric object may be referred to by an identifier which is
called the \emph{label} of the object. For example, the point
$(x,y)$ with label $P$ is represented as $P:=(x,y)$; the straight
line with label $l$ passing through $P_1$ and $P_2$ is represented
as $l:=\mbox{\texttt{line}}(P_1,P_2)$.

Recognition of a basic geometric object $\mathcal{O}$ means to
determine the parameter values of $\mathcal{O}$. For example, a
circle can be recognized by determining the coordinates of its
center and the value of its radius.

Our approach to recognizing basic geometric objects from images of diagrams is based on Hough transform~\cite{hough}, a general technique
for estimating the parameters of a shape from its boundary points.
Through Hough transform, the detection of a shape is converted to a
voting procedure carried out in a parameter space. For instance, the
detection of a line can be realized by checking whether the number
of curves (corresponding to the points on the line) crossing at a
certain point (corresponding to the line) in the parameter space is
greater than a threshold. However, due to the effects of image
quality, line width, and concrete recognition requirements, the
results obtained via Hough transform might not perfectly reflect the
actual features of the geometric diagrams. For example, a line may
be detected as several disconnected short segments; the position and
the size of a circle may be not the same as they are in the diagram.
To improve the accuracy of recognition for basic geometric objects, we
adopt some techniques to refine the results of Hough transform, as
described in the following algorithm.\footnote{To allow use of
previously retrieved information, recognition tasks are arranged in
the order of circles, lines, and then points.}

\begin{algorithm}[Geometric object recognition]\label{algorithm:obj}\rm Given an image $\mathrm{I}$ of diagram, output a set $\mathbb{C}$ of circles, a set $\mathbb{L}$ of lines, and a set $\mathbb{P}$ of points of interest contained in $\mathrm{I}$.

\begin{step}\rm[Recognize circles] \label{reccircle}

\begin{substep}\rm[Preprocess]
Perform graying and smoothing operations on the image $\mathrm{I}$ (using, e.g., the technique of Gaussian smoothing given in~\cite{smoothing}) to obtain a new image $\mathrm{I}_1$.
\end{substep}

\begin{substep}\rm[Detect] Apply the gradient-based Hough transform (see algorithm 21HT  in~\cite{gbht})
 on $\mathrm{I}_1$ to acquire a set $\mathbb{C}$ of circles.
\end{substep}

\begin{substep}\rm[Refine] For each $c:=\mbox{\texttt{circle}}(O,r)$ in $\mathbb{C}$, collect four points $P_1$, $P_2$, $P_3$, and $P_4$ on the left-bound, right-bound, top-bound, and bottom-bound of $c$ respectively,
and then replace $O$ by the centroid of the quadrilateral
$P_1P_2P_3P_4$ and $r$ by the average of the Euclidean distances
$\|O P_1\|$, $\|O P_2\|$, $\|O P_3\|$, and
$\|O P_4\|$.
\end{substep}
\end{step}

\begin{step}\rm[Recognize lines] \label{recline}
There are three possible defects in the lines detected from an image of
diagram by applying Hough transform: (1) a line in the diagram
is detected as some disconnected short segments; (2) the endpoints
of a segment cannot be accurately detected; (3) some nonexisting
segments may be detected on a circle, in particular when the radius
of the circle is large. The following substeps are used to amend the
defects.

\substep \rm[Preprocess]
  Perform binarization and thinning operations on $\mathrm{I}$ (using,
  e.g., the technique in Zhang's parallel thinning algorithm~\cite{thinning})
  to obtain a new image $\mathrm{I}_2$ .

\substep \rm[Detect]
 Apply the progressive probabilistic Hough transform (see~\cite{ppht})\footnote{Both the gradient-based Hough transform and the progressive probabilistic Hough transform are improved versions of Hough transform: the former makes use of local gradients of the image intensity to reduce the computation time and is efficient for detecting circles, and the latter minimizes the amount of computation needed and is reliable for detecting lines. The two transforms have been implemented in OpenCV, an Open source Computer Vision and machine
learning software library~\cite{opencv}, to detect circles and lines respectively.} on $\mathrm{I}_2$ to acquire a set $\mathbb{L}$ of segments.

\substep \rm[Merge segments] For each pair of $\mbox{\texttt{segment}}(P_{1}, P_{2})$ and $\mbox{\texttt{segment}}(P_{3}, P_{4})$ in $\mathbb{L}$, if
$P_{1},P_{2},P_{3}, \mbox{and}~P_{4}$ are collinear and $\min_{1\leq
i\neq j\leq 4}\{\|P_iP_j\|\}<\tau_{l}$ (where
$\tau_{l}$ is a prespecified tolerance), then the pair of
segments is replaced by a new segment $\mbox{\texttt{segment}}(P,Q)$
such that $P,Q \in \{P_{1},P_{2},P_{3},P_{4}\}$ and $\|PQ\| =
\max_{1\leq i,j\leq 4}\{\|P_iP_j\|\}$. \substep \rm[Determine
endpoints] For each $\mbox{\texttt{segment}}(P,Q)$ in $\mathbb{L}$,
if $P$ and $Q$ can be moved to points $P'$ and $Q'$ outwards along
the two directions of the segment respectively as far as there is no
other point detected in $\mathrm{I}_2$, then replace
$\mbox{\texttt{segment}}(P,Q)$ by $\mbox{\texttt{segment}}(P',Q')$.
\substep \rm[Remove nonexisting segments] For each pair of
$\mbox{\texttt{segment}}(P_{1},P_{2})\in \mathbb{L}$ and
$\mbox{\texttt{circle}}(O,r)\in \mathbb{C}$, if
$|\|P_{1}O\|-r|<\tau_c$, $|\|P_{2}O\|-r|<\tau_c$, and
$|\|P_{3}O\|-r|<\tau_c$, where $\tau_c$ is a prespecified tolerance and
$P_{3}$ is the midpoint of $P_{1}$ and $P_{2}$, then remove
$\mbox{\texttt{segment}}(P_{1},P_{2})$ from $\mathbb{L}$.

\substep \rm[Determine types of lines] A point with coordinates
$(x,y)$ is called a \emph{boundary point} of the image
$\mathrm{I}_2$, if $0\leq x< \delta$, or
$W_{\mathrm{I}_2}-\delta\leq x < W_{\mathrm{I}_2}$,
or $0\leq y < \delta$, or $H_{\mathrm{I}_2}-\delta\leq y <
H_{\mathrm{I}_2}$.\footnote{Here $W_{\mathrm{I}_2}$
denotes the width of $ \mathrm{I}_2$,
$H_{\mathrm{I}_2}$ denotes the height of
$\mathrm{I}_2$, and $\delta$ is a given tolerance.} For each
$\mbox{\texttt{segment}}(P,Q)$ in $\mathbb{L}$, if both $P$ and $Q$
are boundary points, then replace $\mbox{\texttt{segment}}(P,Q)$ by
$\mbox{\texttt{line}}(P,Q)$; if $P$ is a boundary point, but $Q$ is
not, then replace $\mbox{\texttt{segment}}(P,Q)$ by
$\mbox{\texttt{halfline}}(P,Q)$; if $Q$ is a boundary point, but $P$
is not, then replace $\mbox{\texttt{segment}}(P,Q)$ by
$\mbox{\texttt{halfline}}(Q,P)$.
\end{step}

\begin{step}\rm[Collect points of interest] \label{colpoints}
The set $\mathbb{P}$ of points of interest are obtained as follows.
\begin{itemize}
\item For each $\mbox{\texttt{circle}}(O,r)$ in $\mathbb{C}$, add $O$ to $\mathbb{P}$;
for each $\mbox{\texttt{circle}}(A,B,C)$ in $\mathbb{C}$, add $A$,
$B$, and $C$ to $\mathbb{P}$.
\item For each $\mbox{\texttt{line}}(P,Q)$, or $\mbox{\texttt{halfline}}(P,Q)$,
or $\mbox{\texttt{segment}}(P,Q)$ in $\mathbb{L}$, add $P$ and $Q$
to $\mathbb{P}$.
\item For each pair of lines in $\mathbb{L}$, compute the numeric coordinates of the intersection point $P$ of the two lines and add $P$ to $\mathbb{P}$, if $P$ exists.
\item For each pair of a line in $\mathbb{L}$ and a circle in $\mathbb{C}$, compute the
numeric coordinates of the intersection points $P_1$ and $P_2$ of
the line and the circle and add $P_1$ and $P_2$ to $\mathbb{P}$, if
$P_1$ and $P_2$ exist.
\item For each pair of circles in $\mathbb{C}$, compute the numeric coordinates of the intersection points $P_1$ and $P_2$ of the two circles and add $P_1$ and $P_2$ to $\mathbb{P}$, if $P_1$ and $P_2$ exist.
\end{itemize}
Due to errors of numeric computation, the same point in the diagram
may be collected into $\mathbb{P}$ more than once with different
coordinates. Therefore, in the above process of adding a point $P$
to $\mathbb{P}$, the following substep need be performed to check
whether $P$ is already contained in $\mathbb{P}$.

\substep\rm[Identify identical points] For any given point $P$, if
there exists a point $P_0$ in $\mathbb{P}$ such that $\|PP_0\|
< \tau_{p}$ (where $\tau_{p}$ is a prespecified tolerance),
then $P$ and $P_0$ are viewed as being identical and $P$ need
not be added to $\mathbb{P}$.
\end{step}
\end{algorithm}

To each of the recognized basic geometric objects, it is necessary
to assign a unique label (or letter), so that geometric relations among the
objects can be expressed clearly. Labels for important geometric
objects (such as points) are usually contained in diagrams. We shall
present a method to extract information on label assignment from
images of diagrams in the next subsection.

\subsection{Recognizing Labels of Geometric Objects}\label{labelrec}

 Labels in an image of diagram may be recognized by checking whether
 each of them matches a character template, as shown in the following
 algorithm.

 \begin{algorithm}[Label recognition]\label{algorithm:label}\rm~Given
 an image $\mathrm{I}$ of diagram and the three sets $\mathbb{C}$, $\mathbb{L}$,
 and $\mathbb{P}$ obtained by applying Algorithm~\ref{algorithm:obj} to $\mathrm{I}$,
 output a list $\mathbf{L}$ of labels in $\mathrm{I}$ and a list $\mathbf{P}$
 of the corresponding centers of the regions where the labels occur in $\mathrm{I}$.

\step\rm[Prepare character templates]
 Produce a predetermined set $\mathbb{T}$ of binary images of letters with font type $\textsf{T}$ commonly used in geometric documents as character templates.

\step\rm[Preprocess]
   Redraw the points in $\mathbb{P}$, the lines in $\mathbb{L}$, and the circles in $\mathbb{C}$ on $\mathrm{I}$ with white (background) color. Perform graying and
   binarization operations on $\mathrm{I}$ to obtain a new image
   $\mathrm{I}_3$. Set $\mathbf{L}:=[~]$ and
   $\mathbf{P}:=[~]$.

\step\rm[Cut out blocks with labels] For each label $L$, use an
alterable rectangular cutting window $W(l,r,t,b)$ to cut out from the image
$\mathrm{I_3}$ a minimal block $\mathrm{B}$ containing the region where $L$ occurs, where $(l,t)$, $(r,t)$, $(l, b)$, and $(r,b)$ denote,
respectively, the left-top, right-top, left-bottom, and right-bottom
vertices of the window. Cutting windows are determined as follows.

Let $H_{\mathrm{I}_3}$ and $W_{\mathrm{I}_3}$ be the height and the
width of $\mathrm{I}_3$, respectively. Set ${\cal
B}$, the set of image blocks with labels, to be empty. For each black (foreground color) pixel point $P$ with coordinates
$(x,y)$ in $\mathrm{I}_3$, but not in any of the cutting
windows for the image blocks in ${\cal B}$, let $h$ and $w$
be initialized to $y$ and $x$, respectively, and do the following.
While $0<h < H_{\mathrm{I}_3}$ and $0< w < W_{\mathrm{I}_3}$ repeat:
    \begin{enumerate}
    \item if the point $(w,h+1)$ is a black pixel point,
    then set $h:=h+1$;
    \item else if the point $(w-1,h+1)$ is a black pixel point,
    then set $w:=w-1$ and $h:=h+1$;
    \item else if the point $(w+1,h+1)$ is a black pixel point,
    then set $w:=w+1$ and $h:=h+1$;
    \item else if the point $(w-1,h)$ is a black pixel point,
    then set $w:=w-1$;
    \item otherwise, break.
    \end{enumerate}
    Then set $\delta_h:=h - y$ and let $l$, $r$, $t$, $b$ be initialized to $x$, $x$, $y$, $y+\delta_h$, respectively. Decrease $l$ by one each time until there is no black pixel point on $\mbox{\texttt{segment}}((l, y), (l,y+\delta_h))$; decrease $r$
    by one each time until there is no black pixel point on $\mbox{\texttt{segment}}((r, y), (r,y+\delta_h))$. In a similar way, $t$ and $b$ can be determined. Finally, use the obtained window $W(l,r,t,b)$ to cut out $\mathrm{B}$ from the
    image\footnote{To ensure successful determination of the size and position of the cutting window,
    we assume that labels have no overlap with geometric objects in the
    image.} and set ${\cal B}:={\cal B}\cup\{\mathrm{B}\}$.

\step\rm[Match character templates] For each image block
$\mathrm{B}$ in ${\cal B}$, if there exists a character
template $\mathrm{T}$ in $\mathbb{T}$ of letter $L$ such that the
similarity of $\mathrm{T}$ and $\mathrm{B}$ is not less than a
threshold (e.g., 90 percent),\footnote{The similarity of two images
is defined as the ratio of the number of pixels at which the two
images have the same binary values to the total number of pixels
after scaling the two images to the same size.} then append $L$ to
the list $\mathbf{L}$, compute the center
$(\frac{l+r}{2},\frac{t+b}{2})$ of the cutting widow and append it
to the list $\mathbf{P}$. Note that the center of the $i$th label in
$\mathbf{L}$ corresponds to the $i$th point in $\mathbf{P}$.
\end{algorithm}

To assign the recognized labels to corresponding geometric objects,
we adopt the following strategies according to the convention that
in geometry, usually a letter in upper case is used to label its
nearest point and a letter in lower case is used to label its
nearest line. For the $i$th label $L_i$ in $\mathbf{L}$, if $L_i$ is
in upper case, then it is assigned to a point $P$ in $\mathbb{P}$
such that for any other point $P'\in \mathbb{P}$, $\|C_iP\| <
\|C_iP'\|$, where $C_i$ is the $i$th point in $\mathbf{P}$;
if $L_i$ is in lower case, then it is assigned to a line $l$ in
$\mathbb{L}$ such that for any other line $l'\in \mathbb{L}$,
$\|C_il\| < \|C_il'\|$.\footnote{The Euclidean
distance from point $P$ to line $l$ is denoted by $\|Pl\|$.}

\begin{figure}[h]
\begin{center}
\includegraphics[width=5cm]{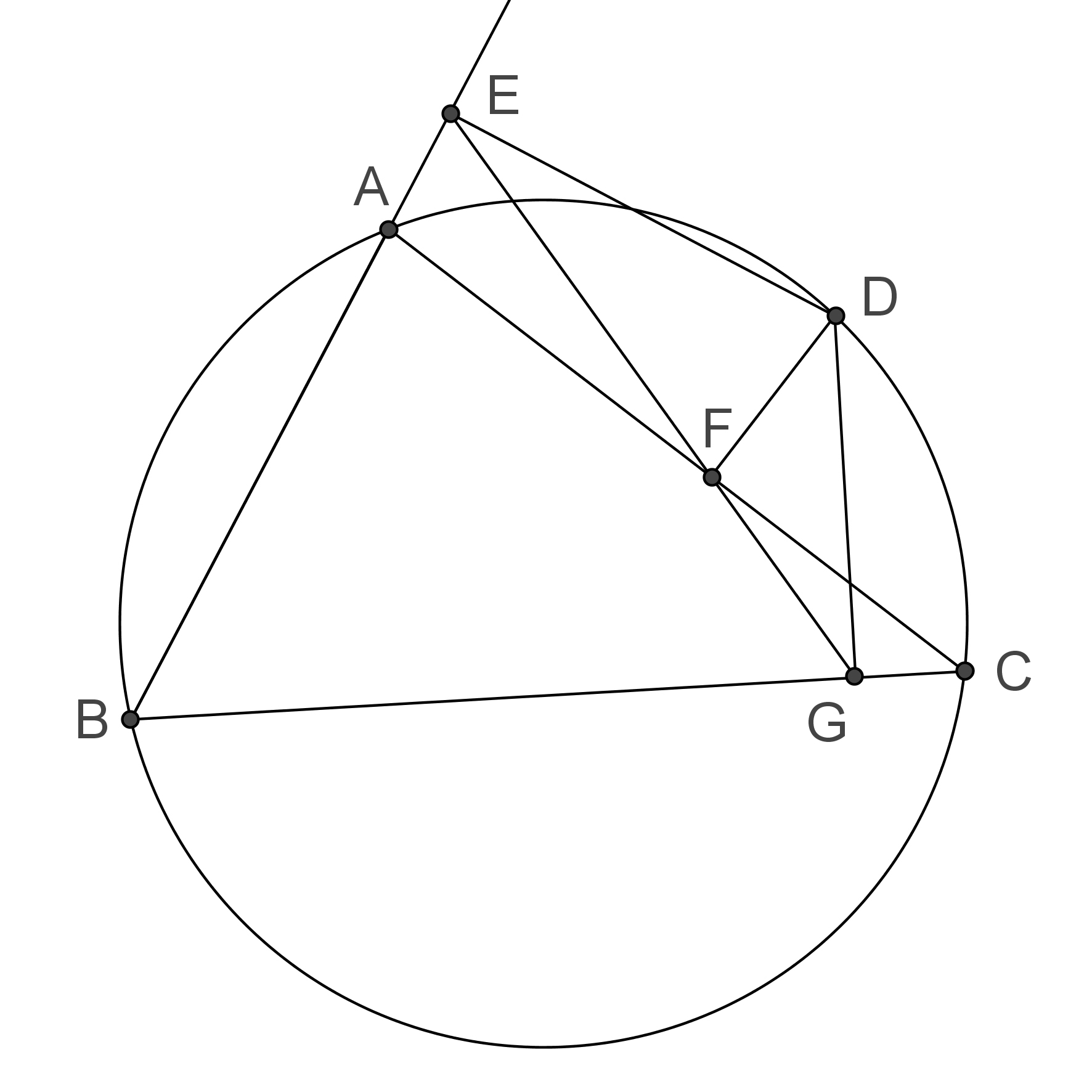}
\caption{An image of diagram for Simson's Theorem}\label{fig:simson}
\end{center}
\end{figure}

For any geometric object that is not labeled in the image, a unique
label is automatically generated by our program to refer to the
object. Taking the image of a diagram (Fig.~\ref{fig:simson}) for
Simson's theorem\footnote{Simson's theorem may be stated as: the
feet of the perpendiculars from a point to the sides of a triangle
are collinear if and only if the point lies on the circumcircle of
the triangle.} as an example, we show the geometric objects obtained
by Algorithm~\ref{algorithm:obj} and the labels recognized by
Algorithm~\ref{algorithm:label} or generated automatically.

\begin{itemize}
\item The set $\mathbb{P}$ of points of interest:
\[\begin{array}{llll}\smallskip
B:=(45,260), & C:=(351,243), & G:=(313,246), & A:=(137,78), \\
\smallskip
F:=(262,174), & H:=(311,212), & E:=(163,37), & I:=(182,0), \\
D:=(305,110), & J:=(196,224), & K:=(184,67), & L:=(224,69).
\end{array}\]
\item The set $\mathbb{L}$ of lines:
\[\begin{array}{lll}\smallskip
a:=\mbox{\texttt{segment}}(B,C), & b:=\mbox{\texttt{segment}}(A,C),
& c:=\mbox{\texttt{segment}}(E, G),\\ \smallskip
d:=\mbox{\texttt{halfline}}(B,I), & e:=\mbox{\texttt{segment}}(E,D), & f:=\mbox{\texttt{segment}}(D,G), \\
g:=\mbox{\texttt{segment}}(F, D).
\end{array}\]
\item The set $\mathbb{C}$ of circle: $h:=\mbox{\texttt{circle}}(J,157)$.
\end{itemize}

The features of diagrams are depicted mainly via geometric relations
(e.g., incidence, perpendicularity, and parallelism) among the
involved objects. Geometric relations play a fundamental role in the
specification of geometric knowledge (e.g., theorems). Based on
retrieved information about geometric objects, we shall present a
method to mine geometric relations in the next subsection.

\subsection{Mining Basic Geometric Relations}
\label{relationsrec}

Some geometric relations such as those listed in
Table~\ref{tablerelations} may be taken as basic geometric relations
because they can be used to describe most features about the size
and position of geometric objects and from them many other geometric
relations can be derived. For example, if point $A$ is incident to
line $p$ and also to line $q$, then the two relations derive the new
relation that $A$ is the intersection point of the two lines $p$ and
$q$.

\begin{table}[h!]
\newcommand{\tabincell}[2]{\begin{tabular}{@{}#1@{}}#2\end{tabular}}
\caption{Basic geometric relations}\label{tablerelations}
\begin{tabular}{p{1.4cm}|p{9cm}|p{3.5cm}}
\hline
\textbf{Type} & \textbf{Representation} & \textbf{Meaning} \\
\hline
onLine & {$\mbox{\texttt{incident}}(A,\!l)$} & point $A$ lies on straight line $l$, or segment $l$, or half line $l$ \\
\hline
onCircle~ & {$\mbox{\texttt{pointOnC}}(A,\!o)$} & point $A$ is on circle $o$ \\
\hline
Parallel & {\tabincell{l}{$\mbox{\texttt{parallel}}(l_1,\!l_2)$}} & $l_1$ is parallel to $l_2$\\
\hline
Perp & {\tabincell{l}{$\mbox{\texttt{perpendicular}}(l_1,\! l_2)$}} & $l_1$ is perpendicular to $l_2$ \\
\hline
dEqual & {\tabincell{l}{$\mbox{\texttt{equal}}(\mbox{\texttt{distance}}(A,\!B), \mbox{\texttt{distance}}(C,\!D))$}} ~~~~~~~~~~~~~~~~~~~~~or $\|AB\|\! = \!\|CD\|$ & the Euclidean distance
 between $A$ and $B$ is equal to that between $C$ and $D$ \\
\hline
aEqual & {\tabincell{l}{$\mbox{\texttt{equal}}(\mbox{\texttt{size}}(\mbox{\texttt{angle}}(A,\!B,\!C)),\! \mbox{\texttt{size}}(\mbox{\texttt{angle}}(D,\!E,\!F)))$}} or $\angle ABC\!=\!\angle DEF$ & the size of $\angle ABC$ is equal to the size of $\angle DEF$ \\
\hline
\end{tabular}
\end{table}

Each basic geometric relation in Table~\ref{tablerelations}
corresponds to an algebraic equality in the coordinates of the
involved points and the radii of the involved circles. In general, a
geometric relation can be certificated to be true if and only if its
corresponding equality holds. Take
$\mbox{\texttt{incident}}(C,\mbox{\texttt{line}}(A,B))$ as an
example and let the coordinates of $A$, $B$, and $C$ be $(x_1,y_1)$,
$(x_2,y_2)$, and $(x_3,y_3)$, respectively. To determine whether $C$
lies on line $AB$, one can check whether the value of the expression
$x_1y_2+x_2y_3+x_3y_1-x_1y_3-x_2y_1-x_3y_2$ is equal to $0$.
However, due to recognition and numeric errors, it is not effective
to determine the equality by simply evaluating the expression, in
particular when the slope of the line is large. We adopt some
techniques to mine basic geometric relations as detailed in the
following algorithm.

\begin{algorithm}[Geometric relation mining]\label{algorithm:relation}\rm~Given the set $\mathbb{P}$ of points of interest, the set $\mathbb{L}$ of lines, and the set $\mathbb{C}$ of circles recognized from an image of diagram with labels, output a set $\mathbb{R}$ of basic relations among the geometric objects.

\step\rm[Mine incidence] Set $\mathbb{R}:=\emptyset$.

\substep \rm For each pair of point $P$ in $\mathbb{P}$ and line $l$
in $\mathbb{L}$, if $\|Pl\|$ is less than a prespecified
tolerance $\tau_{pl}$, then add the relation
$\mbox{\texttt{incident}}(P,l)$ to $\mathbb{R}$.\footnote{The
trivial cases
$\mbox{\texttt{incident}}(A,\mbox{\texttt{line}}(A,B))$ and
$\mbox{\texttt{incident}}(B,\mbox{\texttt{line}}(A,B))$ are ruled
out.}

\substep \rm For each pair of point $P$ in $\mathbb{P}$ and
$\mbox{\texttt{circle}}(O,r)$ in $\mathbb{C}$, if
$|\|AO\|-r|$ is less than a prespecified tolerance
$\tau_{pc}$, then add the relation
$\mbox{\texttt{pointOnC}}(P,\mbox{\texttt{circle}}(O,r))$ to
$\mathbb{R}$.

\substep \rm For each pair of point $P$ in $\mathbb{P}$ and
$\mbox{\texttt{circle}}(A,B,C)$ in $\mathbb{C}$, if
$|\|OA\|-\|OP\||$ (where $O$ is the center of the
circle $ABC$) is less than a prespecified tolerance $\tau_{pc}$,
then add the relation
$\mbox{\texttt{pointOnC}}(P,\mbox{\texttt{circle}}(A,B,C))$ to
$\mathbb{R}$.\footnote{The trivial cases
$\mbox{\texttt{pointOnC}}(A,\mbox{\texttt{circle}}(A,B,C))$,
$\mbox{\texttt{pointOnC}}(B,\mbox{\texttt{circle}}(A,B,C))$, and
$\mbox{\texttt{pointOnC}}(C,\mbox{\texttt{circle}}(A,B,C))$ are
ruled out.}

\step\rm[Mine parallelism and perpendicularity] For each pair of
lines $l_1:=\Box(P_{1},P_{2})$ and $l_2:=\Box(P_{3}, P_{4})$ in
$\mathbb{L}$, where $\Box$ can be \texttt{line}, \texttt{halfline},
or \texttt{segment}, compute $\alpha=\angle \overrightarrow{P_1P_2}$ and $\beta=\angle \overrightarrow{P_3P_4}$ (the angles between the $x$-axis and the vectors
$\overrightarrow{P_1P_2}$ and $\overrightarrow{P_3P_4}$, respectively) according to the following formula:
\[
\angle \overrightarrow{AB}=\left\{ \begin{array}{ll} \smallskip
\frac{3}{2}\,\pi, & \mbox{\texttt{if }}x_B=x_A, \,y_B > y_A; \\\smallskip
\frac{1}{2}\,\pi, & \mbox{\texttt{if }} x_B=x_A, \,y_B < y_A;\\\smallskip
0, & \mbox{\texttt{if }} x_B>x_A,\, y_B = y_A;\\\smallskip
\pi, & \mbox{\texttt{if }} x_B<x_A, \,y_B = y_A;\\\smallskip
 |\arctan(k)|, & \mbox{\texttt{if }}y_B > y_A, \,x_B > x_A;  \\\smallskip
\pi + |\arctan(k)|,~~ & \mbox{\texttt{if }} y_B > y_A, \,x_B< x_A;\\\smallskip
\pi - \arctan(k), & \mbox{\texttt{if }} y_B < y_A, \,x_B< x_A; \\
2\,\pi - \arctan(k), & \mbox{\texttt{if }} y_B < y_A, \,x_B > x_A
 \end{array}\right.
 \]
for any points $A$ and $B$ ($A\neq B$), whose coordinates are $(x_A,y_A)$ and $(x_B,y_B)$ respectively.

Let $\tau_{a}$ be a prespecified tolerance.

\substep \rm If $|\alpha - \beta| < \tau_{a}$ or $||\alpha - \beta|
- \pi |< \tau_{a}$, then add $\mbox{\texttt{parallel}}(l_1,l_2)$ to
$\mathbb{R}$. \substep \rm If $||\alpha - \beta|- \frac{1}{2}\pi| <
\tau_{a}$ or $||\alpha - \beta| - \frac{3}{2}\pi |< \tau_{a}$, then
add $\mbox{\texttt{perpendicular}}(l_1, l_2)$ to $\mathbb{R}$.

\step\rm[Mine distance equality]

    \substep \rm Compute a set $\mathbb{S}$ of segments such that for each $\mbox{\texttt{segment}}(P_1,P_2)\in \mathbb{S}$, $P_1,P_2 \in \mathbb{P}$ and $P_1$ and $P_2$ lie on the same line in
    $\mathbb{L}$.

    \substep \rm For each pair of $\mbox{\texttt{segment}}(A,B)$ and $\mbox{\texttt{segment}}(C,D)$ in $\mathbb{S}$, if $|\|AB\|-\|CD\||$ is less than a prespecified tolerance
    $\tau_{d}$, then add $\mbox{\texttt{equal}}(\mbox{\texttt{distance}}(A,B), \mbox{\texttt{distance}}(C,D))$ to
    $\mathbb{R}$.

\step\rm[Mine angle size equality]

    \substep \rm  Compute a set $\mathbb{P^*}$  of points such that each point in $\mathbb{P^*}$ lies on at least three lines in $\mathbb{L}$.

    \substep \rm For each $P \in \mathbb{P^*}$, compute a list $\mathbb{V}_P$ of vectors such that for each $\overrightarrow{PQ}\in \mathbb{V}_P$,  $Q$ is one of the parameters of the line $l$ in $\mathbb{L}$ and $P$ is incident to $l$. The vectors in $\mathbb{V}_P$ are sorted by the angles between the vectors and the $X$-axis.

    \substep \rm For each $P \in \mathbb{P^*}$, compute a set $\mathbb{A}_P$ of angles such that for each $\angle A_1PA_2\in \mathbb{A}_P$, $\overrightarrow{PA_1},\overrightarrow{PA_2} \in \mathbb{V}_P$ and $A_1\neq A_2$.

    \substep \rm For each pair of $\angle ABC$ and $\angle DEF$ in $\mathbb{A}_P$, if $|\angle ABC - \angle DEF| < \tau_{a}$, then add $\mbox{\texttt{equal}}(\mbox{\texttt{size}}(\mbox{\texttt{angle}}(A,B,C)), \mbox{\texttt{size}}(\mbox{\texttt{angle}}(D,E,F)))$ to $\mathbb{R}$.
\end{algorithm}

Using Algorithm~\ref{algorithm:relation}, one may obtain the
following basic geometric relations for Fig.~\ref{fig:simson}:
\[
\begin{array}{l}\smallskip
\mbox{\texttt{incident}}(G,a),~\mbox{\texttt{incident}}(A,d),~\mbox{\texttt{incident}}(F,b),~\mbox{\texttt{incident}}(F,c),\\
\smallskip
\mbox{\texttt{incident}}(H,f),~\mbox{\texttt{incident}}(E,d),~\mbox{\texttt{incident}}(K,c),~\mbox{\texttt{incident}}(L,e),\\
\smallskip
\mbox{\texttt{incident}}(H,b),~\mbox{\texttt{pointOnC}}(B,h),~\mbox{\texttt{pointOnC}}(C,h),~\mbox{\texttt{pointOnC}}(A,h),\\
\smallskip
\mbox{\texttt{pointOnC}}(K,h),~\mbox{\texttt{pointOnC}}(L,h),~\mbox{\texttt{pointOnC}}(D,h),\\
\mbox{\texttt{perpendicular}}(a,f),~\mbox{\texttt{perpendicular}}(b,g),~\mbox{\texttt{perpendicular}}(d,e).
\end{array}
\]

\section{Automated Generation of Geometric Theorems}\label{theoremDiscovery}

It is remarkable that geometric objects and their relations
retrieved from a single image of diagram allow certain nontrivial
properties implied in the diagram to be expressed explicitly. Such
properties often hold generally for families of diagrams and may be
stated as propositions. A geometric theorem is a true proposition
about the implication of a geometric relation (called the
\emph{conclusion} of the theorem) in all the diagrams that satisfy
the same set of geometric relations (called the \emph{hypothesis} of
the theorem). It is surprising that geometric theorems can be
generated automatically and effectively from the information
retrieved from images of diagrams in three steps: generating candidate
propositions, ruling out false candidates, and proving the obtained
theorems.

\subsection{Generating Candidates}\label{candidateGener}

A candidate proposition is one that is likely a theorem. It can be
generated in a simple way by selecting one (or more) geometric
relation(s) as the conclusion and taking some other relations as the
hypothesis. As there are geometric objects and relations which are
irrelevant to the features of the diagram, it is necessary to remove
such objects and relations for the efficiency of theorem mining from
candidate propositions.

\begin{figure}[h]
\centering
\includegraphics[width=.5\textwidth]{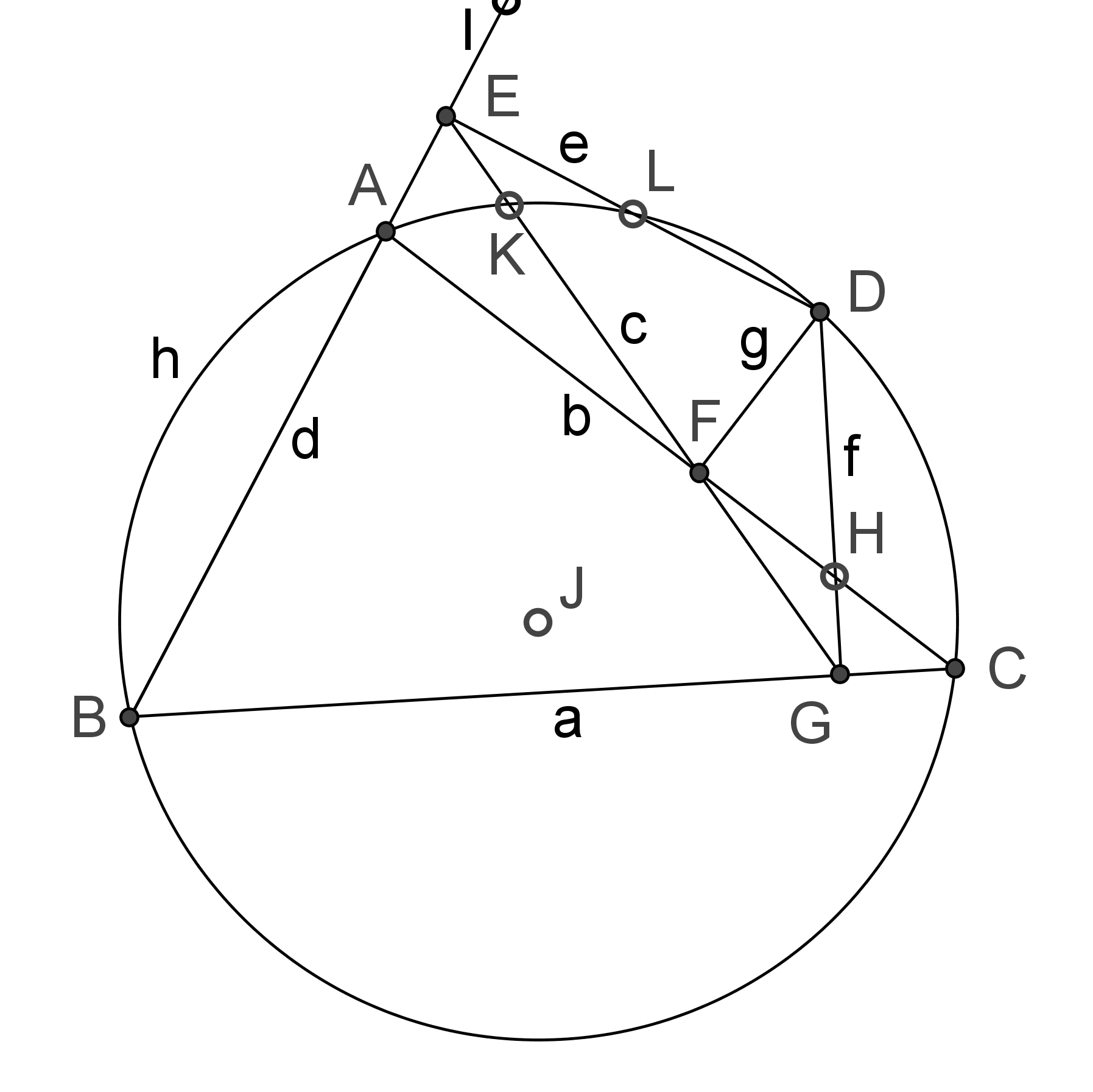}
\caption{Geometric objects recognized from Fig.~\ref{fig:simson}}
\label{simson1}
\end{figure}

Usually more points of interest than needed are recognized from the
diagram. A point of interest is called a point of \emph{attraction}
if it is an endpoint of a segment, or the starting point of a half
line, or the intersection point of two lines, or an intersection
point of two circles or of one line and one circle, or the tangent
point of two circles or of one line and one circle, or an isolated
point. Points of attraction play an important role in forming the
diagram. A point of attraction is called a \emph{characteristic}
point if it is used in the expressions of properties or the
specifications of propositions implied in the diagram. For example,
in the diagram shown in Fig.~\ref{simson1},\footnote{That the point $I$ is truncated purposely in the figure is to show that $I$ is on the boundary of the image.} $I$ is a point of
interest, but not a point of attraction; $K$ $L$, $H$, and $J$ are
points of attraction, but not characteristic points because they are
not used in the specification of Simson's theorem that the diagram
depicts.

A geometric relation is said to be \emph{characteristic} if all the
points used in the relation are characteristic points. To generate
candidate propositions for a diagram we are mainly concerned with
characteristic points and relations. First of all, we introduce the
following rule to remove irrelevant information retrieved.

\begin{ruler}[Remove irrelevant information]\label{rule:irrelativeInfo}\rm
Remove from $\mathbb{P}$ points that are not characteristic and
remove from $\mathbb{R}$ basic geometric relations that are not
characteristic.
\end{ruler}

The following three strategies may be used to implement the above
rule.

\begin{strategy}[Count weights of points]\rm~~~~~~~~~~\\
In general, characteristic points have labels assigned in the
diagram. The more times a point is used in the retrieved geometric
relations, the more likely it is to be characteristic.

To determine which points are potentially characteristic, we weight each point
of interest by the number of its repeating occurrences in the retrieved
relations. Table~2 shows the weights of the points
of interest in Fig.~\ref{simson1} according to the retrieved basic
geometric relations listed in the right column.

\begin{table}[h]
\caption{Weights of the points of interest in Fig.~\ref{simson1}}
\begin{center}
\begin{minipage}[c]{0.4\textwidth}
\begin{tabular}{|c|c|}
\hline
Point of interest & Weight \\ \hline
$B$ & 6 \\ \hline
$C$ & 6 \\ \hline
$G$ & 5 \\ \hline
$A$ & 5 \\ \hline
$F$ & 3 \\ \hline
$H$ & 2 \\ \hline
$E$ & 5 \\ \hline
$I$ & 3 \\ \hline
$D$ & 6 \\ \hline
$J$ & 6 \\ \hline
$K$ & 2 \\ \hline
$L$ & 2 \\ \hline
\end{tabular}
\end{minipage}%
\begin{minipage}[l]{0.6\textwidth}
\begin{tabular}{|c|}
\hline
Basic geometric relations\\
\hline
$\mbox{\texttt{incident}}(G, \mbox{\texttt{segment}}(B,C))$\\ \hline
$\mbox{\texttt{incident}}(A, \mbox{\texttt{halfline}}(B,I))$\\ \hline
$\mbox{\texttt{incident}}(F, \mbox{\texttt{segment}}(A,C))$ \\ \hline
$\mbox{\texttt{incident}}(F, \mbox{\texttt{segment}}(E,G))$ \\ \hline
$\mbox{\texttt{incident}}(H, \mbox{\texttt{segment}}(A,C))$ \\ \hline
$\mbox{\texttt{incident}}(H, \mbox{\texttt{segment}}(D,G))$ \\ \hline
$\mbox{\texttt{incident}}(E, \mbox{\texttt{halfline}}(B,I))$ \\ \hline
$\mbox{\texttt{incident}}(K, \mbox{\texttt{segment}}(E,G))$ \\ \hline
$\mbox{\texttt{incident}}(L, \mbox{\texttt{segment}}(E,D))$ \\ \hline
$\mbox{\texttt{pointOnC}}(B, \mbox{\texttt{circle}}(J,157))$ \\ \hline
$\mbox{\texttt{pointOnC}}(C, \mbox{\texttt{circle}}(J,157))$ \\ \hline
$\mbox{\texttt{pointOnC}}(A, \mbox{\texttt{circle}}(J,157))$ \\ \hline
$\mbox{\texttt{pointOnC}}(D, \mbox{\texttt{circle}}(J,157))$ \\ \hline
$\mbox{\texttt{pointOnC}}(K, \mbox{\texttt{circle}}(J,157))$ \\ \hline
$\mbox{\texttt{pointOnC}}(L, \mbox{\texttt{circle}}(J,157))$ \\ \hline
$\mbox{\texttt{perpendicular}}(\mbox{\texttt{segment}}(B,C), \mbox{\texttt{segment}}(D,G))$ \\ \hline
$\mbox{\texttt{perpendicular}}(\mbox{\texttt{segment}}(A,C), \mbox{\texttt{segment}}(F,D))$\\ \hline
$\mbox{\texttt{perpendicular}}(\mbox{\texttt{halfline}}(B,I), \mbox{\texttt{segment}}(E,D))$ \\ \hline
\end{tabular}
\end{minipage}
\end{center}
\end{table}\label{weightCount}
\end{strategy}

\begin{strategy}[Re-represent lines and circles]\rm~~~~~~~~~~\\
The weights of points of interest depend on the representations of
lines in $\mathbb{L}$ and circles in $\mathbb{C}$, while lines and
circles may be represented in different ways. For example, in
Fig.~\ref{simson1}, the half line $d$ can be represented as
$\mbox{\texttt{halfline}}(B, A)$ or $\mbox{\texttt{halfline}}(B, E)$
instead of $\mbox{\texttt{halfline}}(B,I)$ because $B$, $A$, $E$,
and $I$ are all incident to $d$; the circle can be represented as
$\mbox{\texttt{circle}}(A,B,C)$ or $\mbox{\texttt{circle}}(B,C,D)$
instead of $\mbox{\texttt{circle}}(J,157)$ because $B$, $A$, $D$,
and $C$ are all on the circle. It is therefore desirable to
determine which representation is the best for ruling out the points that are not potentially characteristic. Generally speaking, among the points incident to a line or a circle, the higher weight a point has, the more possible it is to be characteristic. Therefore, we proceed as follows to
re-represent geometric objects according to the weights of points.

\substrategy \rm[Re-represent lines] If $P_1,\ldots,P_n$ are $n$
($\geq 3$) distinct points in $\mathbb{P}$ incident to a straight
line, a segment, or a half line, then the straight line and the
segment are represented as $\mbox{\texttt{line}}(P_i, P_j)$ and
$\mbox{\texttt{segment}}(P_i, P_j)$ respectively, where $P_i$ and
$P_j$ are two distinct points of the highest weights among
$P_1,\ldots,P_n$; the half line is represented as
$\mbox{\texttt{halfline}}(B, P_i)$, where $B$ is the starting point
of the half line and $P_i$ is the point of the highest weight among
$P_1,\ldots,P_n$ and is distinct from $B$.

\substrategy \rm[Re-represent circles] If $P_1,\ldots,P_n$ are $n$
($\geq3$) distinct points in $\mathbb{P}$ and on a circle, then the
circle is represented as $\mbox{\texttt{circle}}(P_i, P_j, P_k)$,
where $P_i$, $P_j$, and $P_k$ are three distinct points of the
highest weights among $P_1,\ldots,P_n$. If there exist other geometric
relations with respect to the center of the circle, add two new
geometric relations
$\mbox{\texttt{equal}}(\mbox{\texttt{distance}}(J,P_i),\mbox{\texttt{distance}}(J,P_j))$ and $\mbox{\texttt{equal}}(\mbox{\texttt{distance}}(J,P_i),\mbox{\texttt{distance}}(J,P_k))$
to $\mathbb{R}$, where $J$ is the center of the circle.

\substrategy \rm[Re-count weights] At each time a geometric object
is re-represented, the weights of the points of interest are
re-counted.

\substrategy \rm[Remove trivial relations] After lines and circles
are re-represented, remove all trivial relations in the form of
$\mbox{\texttt{incident}}(P, \Box(P, *))$,
$\mbox{\texttt{incident}}(P$, $\Box(*,$ $P))$,
$\mbox{\texttt{pointOnC}}(P, \mbox{\texttt{circle}}(P, *, *))$,
$\mbox{\texttt{pointOnC}}(P, \mbox{\texttt{circle}}(*, P, *))$, and
$\mbox{\texttt{pointOnC}}(P, \mbox{\texttt{circle}}(*,
*, P))$ from $\mathbb{R}$ (because they hold obviously), where
$\Box$ can be \texttt{line}, \texttt{segment}, or \texttt{halfline}
and $*$ can be any point in $P_1,\ldots,P_n$.

\end{strategy}

Table~3 shows the weights of the points of interest and basic
geometric relations for Fig.~\ref{simson1} after the
re-representation process.

\begin{center}
\begin{table}[h]
\caption{Re-representations of lines and circles}
\begin{minipage}[t]{0.4\textwidth}
\begin{tabular}{|c|c|}
\hline
Point of interest & Weight \\ \hline
$B$ & 7 \\ \hline
$C$ & 8 \\ \hline
$G$ & 5 \\ \hline
$A$ & 8 \\ \hline
$F$ & 3 \\ \hline
$H$ & 2 \\ \hline
$E$ & 5 \\ \hline
$I$ & 0 \\ \hline
$D$ & 6 \\ \hline
$J$ & 0 \\ \hline
$K$ & 2 \\ \hline
$L$ & 2 \\ \hline
\end{tabular}
\end{minipage}%
\begin{minipage}[t]{0.6\textwidth}
\begin{tabular}{|c|}
\hline
Basic geometric relations \\ \hline
$\mbox{\texttt{incident}}(G, \mbox{\texttt{segment}}(B,C))$\\ \hline
$\mbox{\texttt{incident}}(F, \mbox{\texttt{segment}}(A,C))$ \\ \hline
$\mbox{\texttt{incident}}(F, \mbox{\texttt{segment}}(E,G))$ \\ \hline
$\mbox{\texttt{incident}}(H, \mbox{\texttt{segment}}(A,C))$ \\ \hline
$\mbox{\texttt{incident}}(H, \mbox{\texttt{segment}}(D,G))$ \\ \hline
$\mbox{\texttt{incident}}(E, \mbox{\texttt{halfline}}(B,A))$ \\ \hline
$\mbox{\texttt{incident}}(K, \mbox{\texttt{segment}}(E,G))$ \\ \hline
$\mbox{\texttt{incident}}(L, \mbox{\texttt{segment}}(E,D))$ \\ \hline
$\mbox{\texttt{pointOnC}}(D, \mbox{\texttt{circle}}(A,B,C))$ \\ \hline
$\mbox{\texttt{pointOnC}}(K, \mbox{\texttt{circle}}(A,B,C))$ \\ \hline
$\mbox{\texttt{pointOnC}}(L, \mbox{\texttt{circle}}(A,B,C))$ \\ \hline
$\mbox{\texttt{perpendicular}}(\mbox{\texttt{segment}}(B,C), \mbox{\texttt{segment}}(D,G))$ \\ \hline
$\mbox{\texttt{perpendicular}}(\mbox{\texttt{segment}}(A,C), \mbox{\texttt{segment}}(F,D))$ \\ \hline
$\mbox{\texttt{perpendicular}}(\mbox{\texttt{halfline}}(B,A), \mbox{\texttt{segment}}(E,D))$ \\ \hline
\end{tabular}
\end{minipage}
\end{table}\label{table:rerepresentation}
\end{center}

\begin{strategy}[Determine characteristic points and relations]\rm~~~\\
After geometric objects are re-represented, the points of interest
may be partially determined to be points of attraction or
characteristic points according to their weights as follows.

\substrategy \rm[Determine points of attraction] If the weight of a
point of interest is $0$, then the point is not a point of
attraction because it is not used in any geometric relation.

\substrategy \rm[Determine characteristic points] If the weight of a
point $P$ of interest is less than 3, then $P$ could not potentially be a
characteristic point because it is used at most in two geometric
relations according to the weight counting. This can be explained as follows.
\begin{itemize}
\item If $P$ is the intersection point of two lines or an intersection point of one line and one circle, then no other geometric relations involve $P$ and therefore $P$ could not potentially be a characteristic point.
\item If $P$ is an endpoint of only one segment $l$ or the starting point of only one half line $l$, then no other lines or circles pass through $P$ and there are at most two geometric relations which involve $l$. However, a nontrivial proposition usually needs at least two geometric relations that involve $l$ in the hypothesis. Therefore, in this case, $P$ could not potentially be a characteristic point.
\item If $P$ is the common endpoint of two segments $l_1$ and $l_2$, then only one geometric relation involves $l_1$ and only one geometric relation involves $l_2$. Since a nontrivial proposition usually needs at least two geometric relations that involve the same line in the hypothesis, $P$ could not potentially be a characteristic point in this case.
\end{itemize}
For example, the weights of $H$, $K$, $L$, $J$, and $I$ in Fig.~\ref{simson1} are $2$,
$2$, $2$, $0$, and $0$ respectively as shown in Table~3, so the
points $H$, $K$, $L$, $J$, and $I$ are not characteristic. Together
with the non-characteristic relations, they are removed by
Rule~\ref{rule:irrelativeInfo} (see Table~4).
\begin{table}[h]
\caption{Non-characteristic points and relations removed}
\begin{center}
\begin{minipage}[r]{0.4\textwidth}
\begin{tabular}{|c|c|}
\hline
Characteristic point & Weight \\ \hline
$B$ & 7 \\ \hline
$C$ & 8 \\ \hline
$G$ & 5 \\ \hline
$A$ & 8 \\ \hline
$F$ & 3 \\ \hline
$E$ & 5 \\ \hline
$D$ & 6 \\ \hline
\end{tabular}
\end{minipage}%
\begin{minipage}[l]{0.6\textwidth}
\begin{tabular}{|c|}
\hline
Characteristic relations \\ \hline
$\mbox{\texttt{incident}}(G, \mbox{\texttt{segment}}(B,C))$ \\ \hline
$\mbox{\texttt{incident}}(F, \mbox{\texttt{segment}}(A,C))$ \\ \hline
$\mbox{\texttt{incident}}(F, \mbox{\texttt{segment}}(E,G))$ \\ \hline
$\mbox{\texttt{incident}}(E, \mbox{\texttt{halfline}}(B,A))$ \\ \hline
$\mbox{\texttt{pointOnC}}(D, \mbox{\texttt{circle}}(A,B,C))$ \\ \hline
$\mbox{\texttt{perpendicular}}(\mbox{\texttt{segment}}(B,C), \mbox{\texttt{segment}}(D,G))$ \\ \hline
$\mbox{\texttt{perpendicular}}(\mbox{\texttt{segment}}(A,C), \mbox{\texttt{segment}}(F,D))$ \\ \hline
$\mbox{\texttt{perpendicular}}(\mbox{\texttt{halfline}}(B,A), \mbox{\texttt{segment}}(E,D))$ \\ \hline
\end{tabular}
\end{minipage}
\end{center}
\end{table}\label{table:removeIrrelative}
\end{strategy}

Some of the geometric relations in $\mathbb{R}$ may be derivable
from other relations in $\mathbb{R}$. We call geometric relations
$D_1,\ldots,D_s$ ($1\leq s$) \emph{branch} relations with respect to
other geometric relations $H_1,\ldots,H_d$ ($1\leq d$) if
$D_1,\ldots,D_s$ can be easily derived from $H_1,\ldots,H_d$ on a
sub-diagram. The formula in the form of $H_1,\ldots,H_d\Rightarrow
D_1,\ldots,D_s$ is used to represent that the branch relations
$D_1,\ldots,D_s$ are obtained from $H_1,\ldots,H_d$.

\begin{figure}[h]
\begin{center}
\begin{minipage}[c]{0.5\textwidth}
\centering\includegraphics[width=4.5cm]{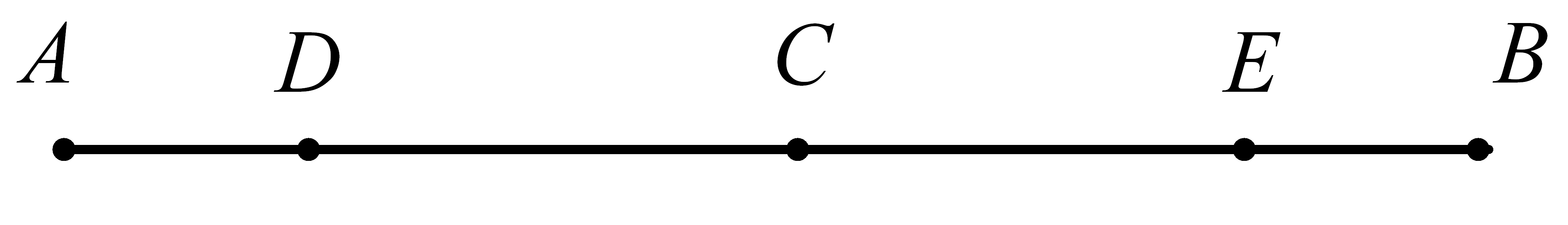}
\caption{Distance relations on the same line} \label{fig:redcon1}
\end{minipage}%
\begin{minipage}[c]{0.5\textwidth}
\centering\includegraphics[width=4.5cm]{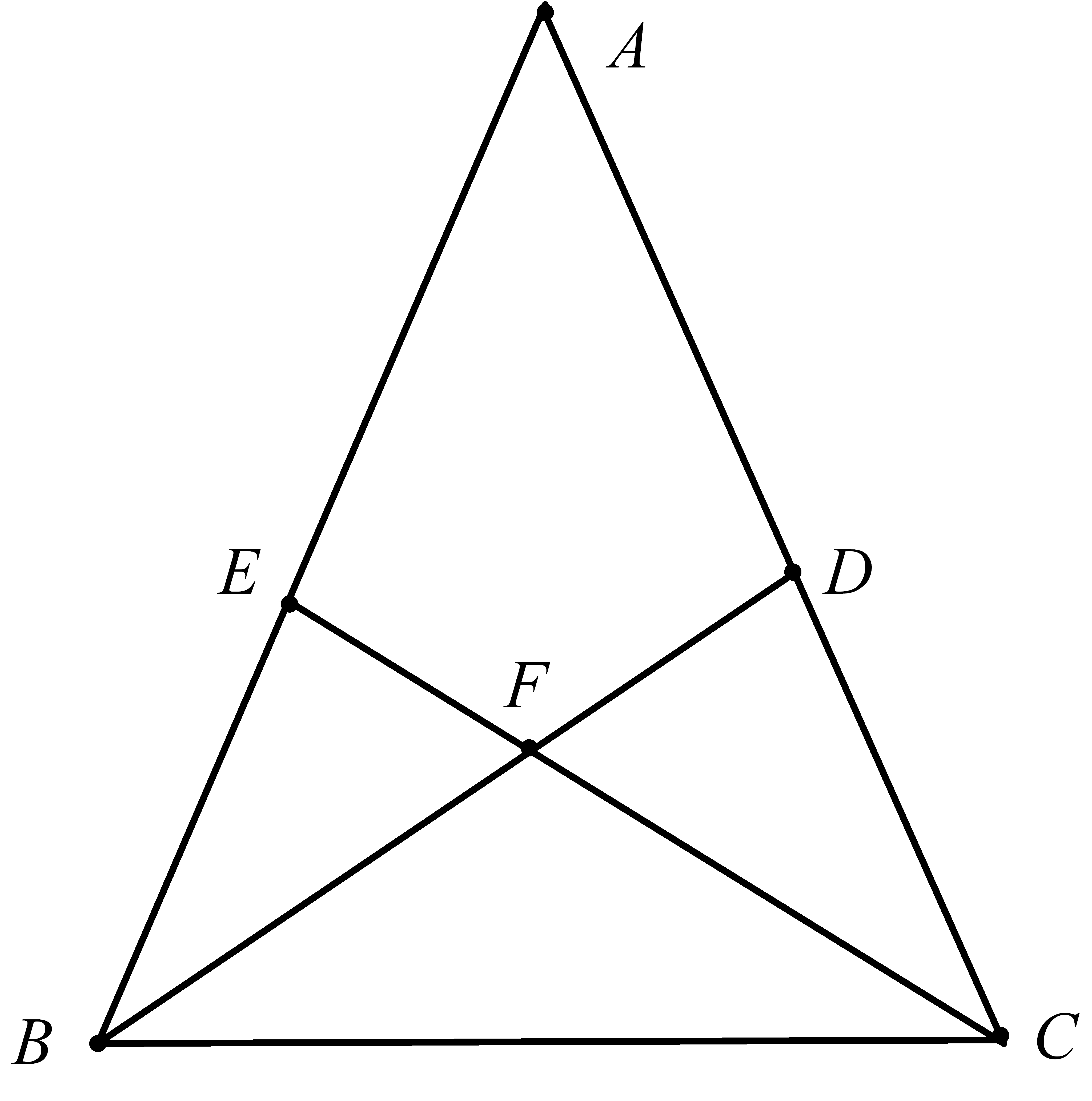}
\caption{Distance relations on different lines } \label{fig:redcon2}
\end{minipage}
\end{center}
\end{figure}

In Fig.~\ref{fig:redcon1} (a sub-diagram for Butterfly theorem), $C$
is the midpoint of segment $AB$ and segment $DE$. Then the following
four relations can be obtained: (1) $\|AC\|=\|CB\|$; (2) $\|DC\|=\|CE\|$; (3)
$\|AD\|=\|EB\|$; (4) $\|AE\|=\|DB\|$. It is easy to see that $(1),(2)\Rightarrow
(3),(4)$; $(1),(3)\Rightarrow (2),(4)$; $(1),(4)\Rightarrow
(2),(3)$.

Similarly, in Fig.~\ref{fig:redcon2} (a diagram for Steiner's
theorem), segment $BD$ and segment $CE$ are internal bisectors of
$\angle ABC$ and $\angle BCA$ respectively and $F$ is the
intersection point of $BD$ and $CE$. Then the following relations
can be obtained: (1) $\|AB\|=\|AC\|$; (2) $\|BD\|=\|CE\|$; (3) $\|AE\|=\|AD\|$; (4) $\|BE\|=\|CD\|$.
One sees that $(1),(2)\Rightarrow (3),(4)$; $(1),(3)\Rightarrow
(2),(4)$; $(1),(4)\Rightarrow (2),(3)$.

Branch relations are usually not used in theorems about the features
of the whole diagram. Therefore, we introduce the following rule.

\begin{ruler}[Remove branch relations]\rm
If $R_1,\ldots,R_k \Rightarrow R_{k+1},\ldots,R_m$ ($1\leq k <m$ and $R_{1},\ldots,R_m$ are all basic geometric relations in $\mathbb{R}$), then
remove $R_{k+1},\ldots,R_m$ from $\mathbb{R}$.
\end{ruler}

There may be different branch relations in the same set of geometric
relations (see, e.g., Figs.~\ref{fig:redcon1}
and~\ref{fig:redcon2}). The following strategy may be used to select
appropriate branch relations.

\begin{strategy}[Determine branch relations] \rm
 Let $\mathbb{\mathbb{E}}$ be the set of distance relations in $\mathbb{R}$.

\substrategy\rm In the case when all the points used in distance
relations lie on the same line (see Fig.~\ref{fig:redcon1}), sort
the points in ascending order of $x$-coordinate or $y$-coordinate to
obtain a list $[P_1,\ldots,P_n]$. Then branch relations are
obtained from the following formulae:
    \begin{enumerate}\smallskip
    \item $\|P_iP_{j}\| = \|P_kP_{l}\| \Rightarrow \|P_iP_k| = \|P_{j}P_{l}\|$;

\smallskip
    \item $\|P_iP_m\|=\|P_mP_l\|, \|P_jP_m\|=\|P_mP_k\| \Rightarrow
\|P_iP_j\|=\|P_kP_l\|,\|P_iP_k\|=\|P_jP_l\|$,\footnote{A distance relation of the
form $\|A_1A_2\|=\|A_3A_4\|$ can be replaced by $\|A_1A_2\|=\|A_4A_3\|$,
$\|A_2A_1\|=\|A_3A_4\|$, or $\|A_2A_1\|=\|A_4A_3\|$.}
    \end{enumerate}
    where $1\leq i<j<m<k<l\leq n$.

\substrategy\rm In the case when some points used in distance
relations lie on different lines (see Fig.~\ref{fig:redcon2}), form
a new set $\mathbb{E}_1$ of distance equations from $\mathbb{E}$
such that the used points are parameters of some retrieved lines.
Then branch relations are obtained from the formula
\begin{equation}
\begin{array}{c}
\|AB\|=\|CD\|, \|AE\|=\|FD\| \Rightarrow \|BE\|=\|CF\|
\end{array}\label{chuan}
\end{equation}
such that $\|AB\|=\|CD\|$ and $\|AE\|=\|FD\|$ are both in $\mathbb{E}_1$, but
$\|BE\|=\|CF\|$ is not in $\mathbb{E}_1$. Let $\mathbb{E}_2$ be the set of
all the obtained branch relations. For each pair of $E_1\in
\mathbb{E}_1$ and $E_2\in \mathbb{E}_2$, a new branch relation $E_3$
with respect to $E_1$ and $E_2$ is determined also by the formula
(1) if $E_3$ is not in $\mathbb{E}_1$.

\substrategy\rm In particular, the distance between the center of a
circle and any point on the circle is constant. Therefore,
branch relations can be determined from the formula
\[\mbox{\texttt{pointOnC}}(A,\mbox{\texttt{circle}}(O,r)), \mbox{\texttt{pointOnC}}(B,\mbox{\texttt{circle}}(O,r))\Rightarrow \|OA\|=\|OB\|.\]
\end{strategy}

As discussed in Section~\ref{relationsrec}, retrieved geometric
relations in $\mathbb{R}$ are basic and from them other new
geometric relations can be derived. A geometric relation $R$ is
called a \emph{derived} relation if it is implied by the
basic geometric relations $R_1,\ldots,R_m$~$(2\leq m)$. The formula
$R_1,\ldots,R_m\mapsto R$ is used to represent that $R$ can be
obtained from $R_1,\ldots,R_m$.

\begin{ruler}[Introduce derived relations]\rm
If $R_1,\ldots,R_m\mapsto R$ and for all $1\leq i \leq m$, $R_i\in
\mathbb{R}$ and $R\notin \mathbb{R}$, then remove $R_1,\ldots,R_m$ from $\mathbb{R}$ and add
$R$ to $\mathbb{R}$.
\end{ruler}

\begin{strategy}[Introduce new geometric objects]\rm~\\
Derived relations can be obtained from the following formulae:
\begin{enumerate}\smallskip
\item $\mbox{\texttt{incident}}(C,p), \mbox{\texttt{incident}}(C,q) \mapsto C := \mbox{\texttt{intersection}}(p,q)$;\footnote{A geometric
relation of the form $L:=f(p_1,p_2,\ldots,p_n)$ means that $L$ is
the label for the geometric object $f(p_1,p_2,\ldots,p_n)$ and
$\mbox{\texttt{intersection}}(p,q)$ denotes the intersection point
of $p$ and $q$.}

\smallskip
\item $\mbox{\texttt{incident}}(C,\mbox{\texttt{segment}}(A,B)), AC = CB \mapsto C := \mbox{\texttt{midpoint}}(A, B)$;\footnote{$\mbox{\texttt{midpoint}}(A, B)$ denotes the midpoint of segment
$AB$.}

\smallskip
\item $\mbox{\texttt{incident}}(C,p), \mbox{\texttt{perpendicular}}(p,q) \mapsto C := \mbox{\texttt{foot}}(p, q)$.\footnote{$\mbox{\texttt{perpendicular}}(p,q)$ can be replaced by $\mbox{\texttt{perpendicular}}(q,p)$ and $\mbox{\texttt{foot}}(p, q)$ denotes the foot of two lines $p$ and $q$ perpendicular to each other.}
\end{enumerate}

For example, from the relations in Table~4 and by Rule~3 one can
obtain the derived geometric relations listed in Table~5.

\begin{table}[h]
\caption{Characteristic relations derived from relations in Table~4}
\center
\begin{minipage}[r]{0.5\textwidth}
\end{minipage}%
\begin{minipage}{0.5\textwidth}
\begin{tabular}{|c|}
\hline
Characteristic relations\\ \hline
$G:=\mbox{\texttt{foot}}(\mbox{\texttt{segment}}(B,C),\mbox{\texttt{segment}}(D,G))$\\ \hline
$F:=\mbox{\texttt{foot}}(\mbox{\texttt{segment}}(A,C),\mbox{\texttt{segment}}(D,F))$\\ \hline
$E:=\mbox{\texttt{foot}}(\mbox{\texttt{halfline}}(B,A),\mbox{\texttt{segment}}(D,E))$\\ \hline
$\mbox{\texttt{incident}}(F, \mbox{\texttt{halfline}}(E,G))$ \\ \hline
$\mbox{\texttt{pointOnC}}(D, \mbox{\texttt{circle}}(A,B,C))$ \\ \hline
\end{tabular}
\end{minipage}
\end{table}\label{upgrading}
\end{strategy}

\begin{strategy}[Generate candidate propositions]\rm~\\
To formulate a proposition, one needs to determine which geometric relations can be taken as the hypothesis, in which order the relations are introduced in the hypothesis, and which one can be taken as the conclusion.

We first introduce an order $\prec$ on the characteristic points
$P_1,\ldots,P_n$ according to the following two rules.
\begin{enumerate}
\item If $P_i$ is the label for a derived relation, then $P_j\prec P_i$, where $P_j$ is used in the relation and $P_j\neq P_i$.
\item Otherwise, if the weight of $P_i$ is higher than that of $P_j$, then $P_i\prec P_j$.
\end{enumerate}

For example, according to the weights of the characteristic points
in Table~4 and the representations of the characteristic relations
in Table~\ref{upgrading}, an order of the characteristic points is
$C\prec A\prec B\prec D \prec G\prec E \prec F$.

Based on the order $\prec$ of points, an order $\lessdot$ is induced
on characteristic relations (after the above-stated rules have been
applied)
\begin{equation*}\label{eq1}
R_1[P_{11},\ldots,P_{1{k_1}}],\,
R_2[P_{21},\ldots,P_{2{k_2}}],\, \ldots,\,
R_{m}[P_{m1},\ldots,P_{m{k_m}}],
\end{equation*}
where $P_{i1},\ldots,P_{i{k_i}}$ are $k_i$ points used in
$R_i~(1\leq i \leq m)$ such that $P_{i1}\prec \cdots \prec
P_{i{k_i}}$, according to the following three rules.

\begin{enumerate}
\item If $P_{i{1}}\prec P_{j{1}}$, then $R_i[P_{i1},\ldots,P_{i{k_i}}]\lessdot R_j[P_{j1},\ldots,P_{j{k_j}}]$.
\item If there exists a $w$ ($1\leq w < \min\{k_i,k_j\}$) such that for all $t$ ($1\leq t \leq w$) $P_{i{t}}$ is identical to $P_{j{t}}$ and $P_{i(w+1)}\prec P_{j(w+1)}$, then $R_i[P_{i1},\ldots,P_{i{k_i}}]\lessdot R_j[P_{j1},\ldots,P_{j{k_j}}]$.
\item Suppose that $k_i \leq k_j$. If for all $t$ ($1\leq t \leq k_i$) $P_{i{t}}$ is identical to $P_{j{t}}$, then  $R_i[P_{i1},\ldots,P_{i{k_i}}]\lessdot R_j[P_{j1},\ldots,P_{j{k_j}}]$.
\end{enumerate}

The characteristic relations listed in Table~\ref{upgrading} are
ordered by $\lessdot$ as: $\mbox{\texttt{incident}}(D,$
$\mbox{\texttt{circle}}(A,B,C))$ $\lessdot$
$F:=\mbox{\texttt{foot}}(\mbox{\texttt{segment}}(A,C),\mbox{\texttt{segment}}(D,F))$
$\lessdot$
$G:=\mbox{\texttt{foot}}(\mbox{\texttt{segment}}(B,C),\mbox{\texttt{segment}}(D,G))$
$\lessdot$ $E:=\mbox{\texttt{foot}}(\mbox{\texttt{halfline}}(B,A),$
$\mbox{\texttt{segment}}(D,E))$ $\lessdot$
$\mbox{\texttt{incident}}(F,$ $\mbox{\texttt{segment}}(E,G))$.

Given $R_1\lessdot R_2\lessdot \cdots \lessdot R_m$, the hypothesis
and conclusion of a candidate proposition are generated according to
the following three rules.
\begin{enumerate}
\item Any basic relation $R_i$ can be taken as the conclusion.
\item If $R_i$ and $R_j$ are both derived relations with the same label, then either $R_i$ or $R_j$ can be taken as the conclusion.
\item The geometric relations other than the conclusion may be taken as the hypothesis.
\end{enumerate}
The generated candidate propositions may be represented in the
following form: Proposition($T_k$,
[$R_1,\ldots,R_{k-1},R_{k+1},\ldots,R_m$],\ [$R_k$]), where $1\leq
k\leq m$, $T_k$ is the name, $R_1,\ldots,R_{k-1},R_{k+1},\ldots,R_m$
the hypothesis, and $R_k$ the conclusion of the proposition.

For example, two candidate propositions may be generated for the
diagram in Fig.~\ref{fig:simson}:
\[
\begin{aligned}
\mbox{Proposition}(\mbox{Simson}_5, [\mbox{\texttt{incident}}(D,\mbox{\texttt{circle}}(A,B,C)), F:=\mbox{\texttt{foot}}(\mbox{\texttt{halfline}}(A,\\
C),\mbox{\texttt{segment}}(D,F)), G:=\mbox{\texttt{foot}}(\mbox{\texttt{segment}}(B,C),\mbox{\texttt{segment}}(D,G)), E:= \\
 \mbox{\texttt{foot}}(\mbox{\texttt{segment}}(B,A),\mbox{\texttt{segment}}(D,E))], [\mbox{\texttt{incident}}(F,\mbox{\texttt{segment}}(E,G))])
\end{aligned}
\] and
\[
\begin{aligned}
\mbox{Proposition}(\mbox{Simson}_1, [F:=\mbox{\texttt{foot}}(\mbox{\texttt{halfline}}(A,C),\mbox{\texttt{segment}}(D,F)), G:=\mbox{\texttt{foot}}\\ (\mbox{\texttt{segment}}(B,C),\mbox{\texttt{segment}}(D,G)), E:=\mbox{\texttt{foot}}(\mbox{\texttt{segment}}(B,A),\mbox{\texttt{segment}}(D,\\
E)),\mbox{\texttt{incident}}(F,\mbox{\texttt{segment}}(E,G))], [\mbox{\texttt{incident}}(D,\mbox{\texttt{circle}}(A,B,C))]).
\end{aligned}
\]

\end{strategy}

\subsection{Ruling out False Candidates}\label{theoremMine}

To verify the truth of a candidate proposition, we use algebraic
methods which have been successfully applied to automated geometric
theorem proving. For the efficiency of theorem mining, false
propositions need be ruled out first, so that each proposition
submitted to a theorem prover is a potential theorem.

A counterexample of a proposition is a diagram for which the
hypothesis of the proposition holds, but the conclusion of the
proposition does not. If a counterexample can be found, then the
proposition must not be a theorem. In what follows we present a
numeric verification technique, based on the characteristic set
method of Wu~\cite{wu,wang}, for finding counterexamples to rule out
false propositions.

\begin{algorithm}[Proposition verification]\label{algorithm:mine}\rm
Given a set $\mathfrak{P}=\{\mathcal{P}_1,\ldots,\mathcal{P}_l\}$ of
candidate propositions, output a set $\mathfrak{F}$ of propositions
that cannot be theorems.

\medskip Set $\mathfrak{F}:=\emptyset$.

For each candidate proposition $\mathcal{P}_t~(1\leq t \leq l)$, do
the following steps.

\begin{step}\rm[Algebraization and triangularization]\label{triangulize}

\substep\rm Assign coordinates $x_j~(1\leq j \leq h)$ (manually or
automatically) to the points used in the hypothesis of
$\mathcal{P}_t$. \substep\rm Translate the geometric relations
$R_1,\ldots,R_{k-1},R_{k+1},\ldots,R_m$ in the hypothesis into
algebraic equations
\begin{equation*}\label{eq1}
\left\{
\begin{array}{rl}
f_1(x_1,\ldots,x_h)&=0,\\
f_2(x_1,\ldots,x_h)&=0,\\
\cdots\cdots\quad\\
f_{m-1}(x_1,\ldots,x_h)&=0,
\end{array}
\right.
\end{equation*}
and the conclusion $R_k$ into an algebraic equation $C=0$. Fix a variable ordering, say $x_1\prec \cdots \prec x_h$, which is either given or chosen heuristically.

\substep\rm Let $\mathbf{P}=\{f_1,\ldots,f_{m-1}\}$ and ${\rm
Zero}(\mathbf{P})$ denote the set of all common zeros of
$f_1,\ldots,f_{m-1}$. Using Wu-Ritt's algorithm, one can compute a
Wu characteristic set $\mathbf{C}$ of $\mathbf{P}$, which has the
following triangular form
\begin{equation*}\label{eqc}
\left[
    \begin{array}{l}
    c_1(x_1,\ldots,x_{p_1}),\\
    c_2(x_1,\ldots,x_{p_1},\ldots,x_{p_2}),\\
    \qquad\qquad\cdots\cdots\\
    c_r(x_1,\ldots,x_{p_1},\ldots,x_{p_2},\ldots,x_{p_r})                \end{array}
\right],
\end{equation*}
such that ${\rm Zero}(\mathbf{P}/I)={\rm Zero}(\mathbf{C}/I)$, where
$I$ is the product of the leading coefficients of the polynomials in
$\mathbf{C}$ with respect to their leading variables, and ${\rm
Zero}(\mathbf{P}/I)={\rm Zero}(\mathbf{P}) \setminus {\rm
Zero}(\{I\})$. If $\mathbf{C}$ consists of a single nonzero
constant, then the geometric relations in the hypothesis are
inconsistent. In this case, add $\mathcal{P}_t$ to $\mathfrak{F}$
and proceed to deal with $\mathcal{P}_{t+1}$; otherwise, go to the
next step.
\end{step}

\begin{step}\rm[Instantiating and solving]\label{solve}
Let
$\mathbf{u}=\{x_1,\ldots,x_h\}\setminus\{x_{p_1},\ldots,x_{p_r}\}$.
Randomly choose a set $\bar{\mathbf{u}}$ of numeric values for the
coordinates in $\mathbf{u}$ and determine (all possible) values
$\bar{x}_{p_1},\ldots,\bar{x}_{p_r}$ for the other coordinates
$x_{p_1},\ldots,x_{p_r}$ by solving the equations
\[c_j|_{\mathbf{u}=\bar{\mathbf{u}},x_{p_1}=\bar{x}_{p_1},\ldots,
x_{p_{j-1}}=\bar{x}_{p_{j-1}}}=0,\quad j=1,\ldots,r,\] successively
for $x_{p_1},\ldots,x_{p_r}$.
\end{step}

\begin{step}\rm[Numeric checking]\label{check}
Compute the numeric value $\bar{C}$ of $C$ at
$\mathbf{u}=\bar{\mathbf{u}}$ and $(x_{p_1},\ldots,x_{p_{r}})
=(\bar{x}_{p_1},\ldots, \bar{x}_{p_{r}})$. If $\bar{C}< \tau_C$ (where $\tau_C$ is a prespecified tolerance determined on the basis of empirical results) for all the solutions $(x_{p_1},\ldots,x_{p_{r}})
=(\bar{x}_{p_1},\ldots, \bar{x}_{p_{r}})$, then the proposition
$\mathcal{P}_t$ is a potential theorem. Otherwise, $\mathcal{P}_t$
must not be a theorem, so it is added to $\mathfrak{F}$.
\end{step}
\end{algorithm}

\subsection{Proving Theorems}\label{theoremProve}

Let $\mathcal{T}_{1},\ldots, \mathcal{T}_{s}$ be the candidate
propositions obtained after ruling out the set $\mathfrak{F}$ of
propositions from $\mathfrak{P}$ by Algorithm~\ref{algorithm:mine}.
Now one can use Wu's method to prove the candidate propositions
automatically.\footnote{Wu's method is complete for proving
geometric theorems involving equalities only.}

For each $\mathcal{T}_d$~($1\leq d \leq s$), let $\mathbf{C}$ be the
Wu characteristic set computed in step~\ref{triangulize} of
Algorithm~\ref{algorithm:mine} with $\mathcal{P}_t=\mathcal{T}_d$.
Then do the following two steps.

\begin{step}\rm[Pseudo-division and irreducible decomposition]
\substep\rm Compute the pseudo-remainder $R$ of the conclusion
polynomial $C$ with respect to $\mathbf{C}$. If $R\equiv 0$, then
${\rm Zero}(\mathbf{C}/I)\subset{\rm Zero}(C)$ and thus under the
condition $I\neq 0$, the proposition $\mathcal{T}_d$ is a theorem.
In this case, go to step 4.5.

\substep\rm Decompose $\mathbf{C}$ into finitely many irreducible
ascending sets $\mathbf{C}_1,\ldots,\mathbf{C}_e$ such that ${\rm
Zero}(\mathbf{P}/I)=\bigcup\nolimits^{e}_{i=1}{\rm
Zero}(\mathbf{C}_i/II_i)$, where each $\mathbf{C}_i$ has the same
triangular form as $\mathbf{C}$ and $I_i$ is the product of the
leading coefficients of the polynomials in $\mathbf{C}_i$ with
respect to their leading variables. Under the condition $I\neq 0$,
each $\mathbf{C}_i$ represents an irreducible component of the
algebraic variety ${\rm Zero}(\mathbf{P})$.

\substep\rm For each $\mathbf{C}_i$ ($1\leq i \leq e$), compute the
pseudo-remainder $R_i$ of $C$ with respect to $\mathbf{C}_i$. If
$R_i\equiv 0$ for some $i$, then under the condition $II_i\neq 0$,
the proposition $\mathcal{T}_d$ is a partially true theorem. If
$R_i\equiv 0$ for all $i$, then under the condition $II_1\cdots
I_e\neq 0$, the proposition $\mathcal{T}_d$ is a theorem.
\end{step}

\begin{step}\rm[Analyzing nondegeneracy conditions]
A candidate proposition may be proved to be a theorem, usually under
certain inequality conditions. Some of the conditions are needed to
ensure that the considered geometric configurations are in generic
position (e.g., a triangle referred to in the proposition does not
degenerate to a line). Such algebraic nondegeneracy conditions may
be translated back into geometric form (see~\cite{geother}). There
are inequality conditions which are not necessarily connected to
nondegeneracy. Those conditions are either unnecessary, or produced
to make a partially true theorem a theorem, or included to make the
statement of the proposition or its algebraic form rigorous.
\end{step}

\section{Implementation and Experiments}\label{implementation}

The effectiveness of the approach we have proposed for automated
generation of geometric theorems from images of diagrams depends on the
completeness and accuracy of the information retrieved as well as
the capability and efficiency of the theorem prover used. In this
section we present some experimental results with a preliminary
implementation of the approach.

The algorithms described in Section~\ref{inforetrieval} have been
implemented in C++. Images of diagrams for testing and character
templates were prepared by using GeoGebra~\cite{geogebra} which is a dynamic
geometry software system for interactive construction of diagrams,
annotation of labels for geometric objects, and exportation of
images. Circles and lines are detected from the images of diagrams by
using functions \emph{cvHoughCircles} and \emph{cvHoughLines2}
provided in OpenCV.

Eight parameters $\tau_l, \tau_c, \delta, \tau_p, \tau_{pl}, \tau_{pc}, \tau_d, \tau_a$ are used to specify tolerances in our approach for retrieving geometric information from images of diagrams. We firstly acquire empirical values $\overline{\tau_l}, \ldots, \overline{\tau_a}$ for $\tau_l, \ldots, \tau_a$ by making experiments on a set of test images with fixed size $400\times400$. Then, for any given image \texttt{I} of diagram, the tolerances will be automatically adjusted according to the size of \texttt{I}. For example, if the size of I is $W\times H$, then $\tau_c$ will be reset to ${\overline{\tau_c}\,\min(W,H)}/{400}$ as $\tau_c$ is used to determine the equality of Euclidean distances. The parameters $\tau_l, \tau_p, \delta, \tau_{pl}, \tau_{pc}, \tau_d$ will be reset similarly, while $\tau_a$ will remain to be $\overline{\tau_a}$ because $\tau_a$ is used to determine the equality of angles which are not affected by image scaling.

The strategies presented in Section~\ref{candidateGener} for
automated generation of candidate propositions have been implemented
in Java. As the computation of characteristic sets and irreducible
triangular decomposition needed in the process of geometric theorem
mining and proving are sophisticated and expensive symbolic
computation processes, we choose to use Epsilon~\cite{epsilon} for
the involved polynomial elimination, triangularization, and
decomposition and GEOTHER~\cite{geother} for automated
algebraization and proof of geometric theorems and automated
interpretation of algebraic nondegeneracy conditions. An interface
for transforming the specifications of candidate propositions into
the native representations of GEOTHER has been developed.

To test our approach, we have made experiments on the images of diagrams
shown in Table~\ref{experiments}.\footnote{The theorems generated
automatically from images of diagrams are presented on the website
http://geo.cc4cm.org/data/recognizer/.} The diagrams used for the experiments were selected from~\cite{shi}, provided that the theorems they illustrate can be expressed by using only the basic geometric relations listed in Table~\ref{tablerelations}. Different diagrams may involve different types of basic relations. For example, the diagrams with Nos.~1 and 2 only involve ``onLine'' relations; the diagrams with Nos.~3, 4, and 5 involve both ``onLine'' and ``dEqual'' relations; the diagram with No.~6 involves both ``onLine'' and ``Perp'' relations. It is easy to figure out from the results of test on the diagrams the capability of the current implementation of our approach. In Table~\ref{experiments}, ``Undesired'' denotes
the number of undesired geometric relations (e.g., those relations
which hold occasionally in the input diagram, but do not hold in
other diagrams for the same theorem); ``Time'' is recorded in seconds for
information retrieval from the image;\footnote{The programs for information retrieval are run on a
machine with 1.86GHz CPU and 1.24G of memory.} ``Candidates''
denotes the number of generated candidate propositions; and ``Theorems''
denotes the number of proved theorems.

\begin{longtable}{c|c|c|c|c|c}
    \caption{Test results}\vspace{-0.3cm}\label{experiments}\\ \hline
    No. &  Image & Undesired & ~Time~ & Candidates & Theorems  \\
    \hline \endfirsthead

    \multicolumn{2}{c}%
    {\tablename\ \thetable{}: Test results} \\ \hline
    No. &  Image & Undesired & ~Time~ & Candidates & Theorems  \\
    \hline \endhead
    1 &
    \begin{minipage}[c]{0.25\textwidth}
        \includegraphics[width=2.8cm]{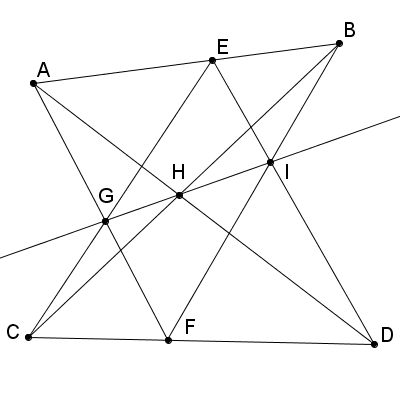}
    \end{minipage} & 2 & 0.25 & 3 & 3\\
    \hline
    2 &
    \begin{minipage}[c]{0.25\textwidth}
         \includegraphics[width=2.8cm]{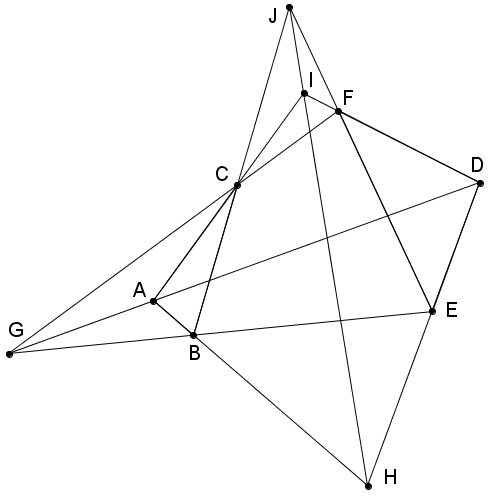}
    \end{minipage} & 9 & 0.25 & 4 & 3 \\
    \hline
    3 &
    \begin{minipage}[c]{0.25\textwidth}
        \includegraphics[width=2.8cm]{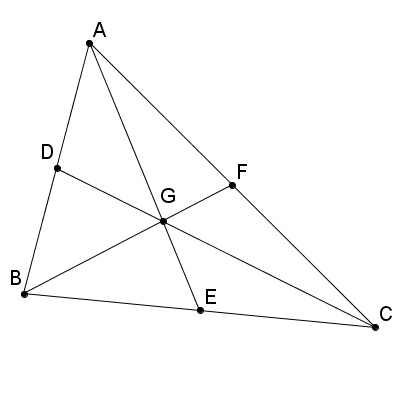}
    \end{minipage} & 0 & 0.187 & 1 & 1 \\
    \hline
    4 &
    \begin{minipage}[c]{0.25\textwidth}
        \includegraphics[width=2.8cm]{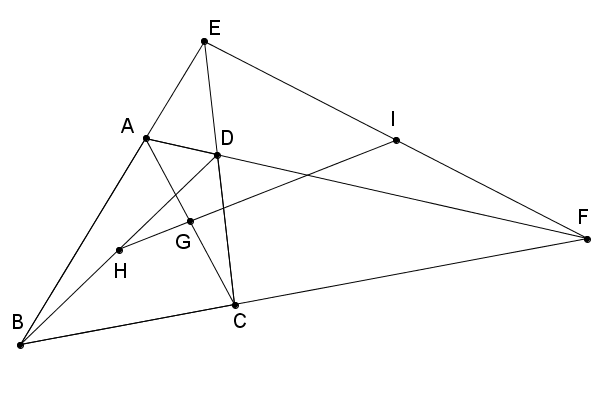}
    \end{minipage} & 7 & 0.203 & 3 & 3\\
    \hline
    5 &
    \begin{minipage}[c]{0.25\textwidth}
         \includegraphics[width=2.8cm]{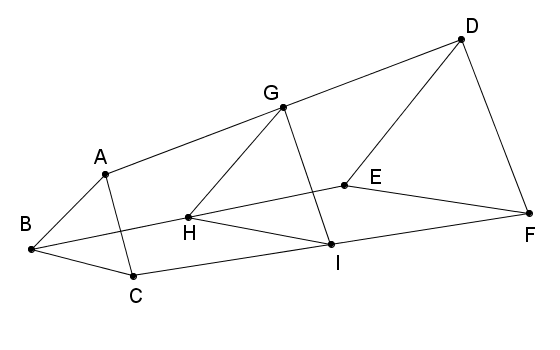}
    \end{minipage} & 3 & 0.203 & 8 & 0 \\
    \hline
    6 &
    \begin{minipage}[c]{0.25\textwidth}
        \includegraphics[width=2.8cm]{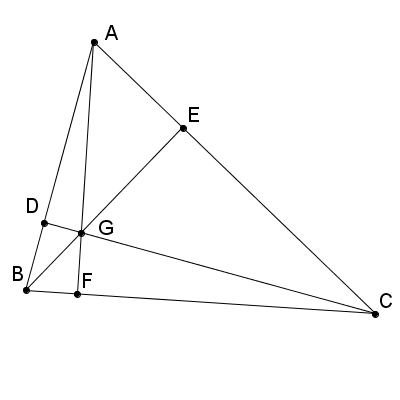}
    \end{minipage} & 0 & 0.14 & 1 & 1 \\
    \hline
    7 &
    \begin{minipage}[c]{0.25\textwidth}
        \includegraphics[width=2.8cm]{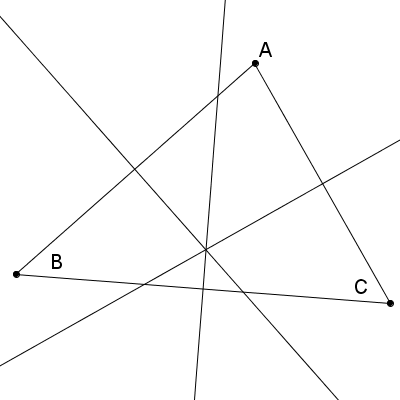}
    \end{minipage} & 0 & 0.156 & 6 & 6 \\
    \hline
    8 &
    \begin{minipage}[c]{0.25\textwidth}
        \includegraphics[width=2.8cm]{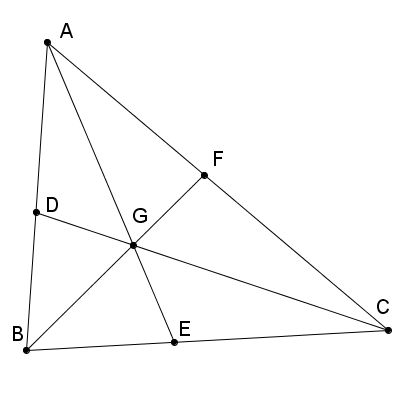}
    \end{minipage} & 0 & 0.187 & 7 & 4 \\
    \hline
    9 &
    \begin{minipage}[c]{0.25\textwidth}
        \includegraphics[width=2.8cm]{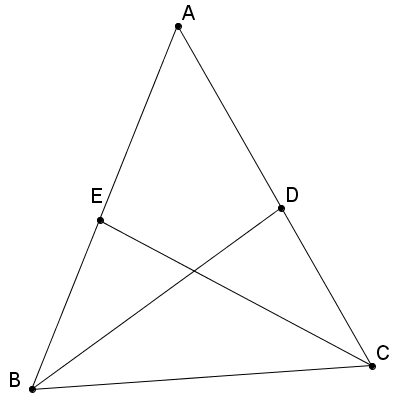}
    \end{minipage} & 0 & 0.124 & 8 & 7 \\
    \hline
    10 &
    \begin{minipage}[c]{0.25\textwidth}
        \includegraphics[width=2.8cm]{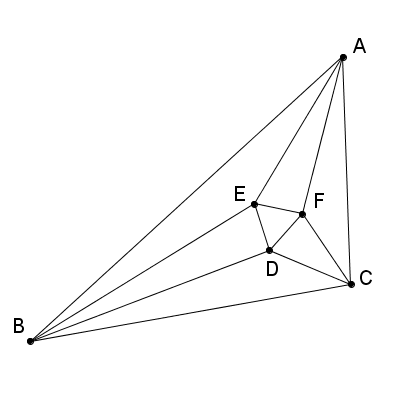}
    \end{minipage} & 0 & 0.187 & 9 & 0\\
    \hline
    9-11 &
    \begin{minipage}[c]{0.25\textwidth}
        \includegraphics[width=2.8cm]{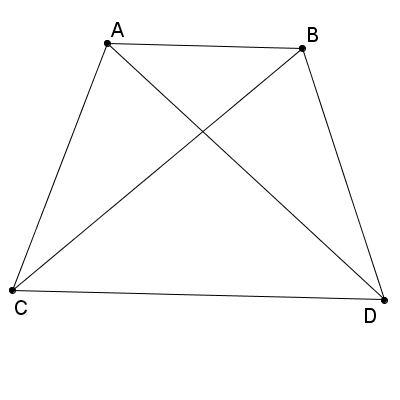}
    \end{minipage} & 0 & 0.14 & 5 & 5\\
    \hline
    12 &
    \begin{minipage}[c]{0.25\textwidth}
         \includegraphics[width=2.8cm]{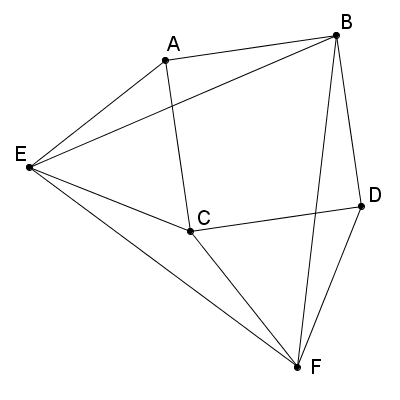}
    \end{minipage} & 0 & 0.156 & 42 & 1 \\
    \hline
    13 &
    \begin{minipage}[c]{0.25\textwidth}
        \includegraphics[width=2.8cm]{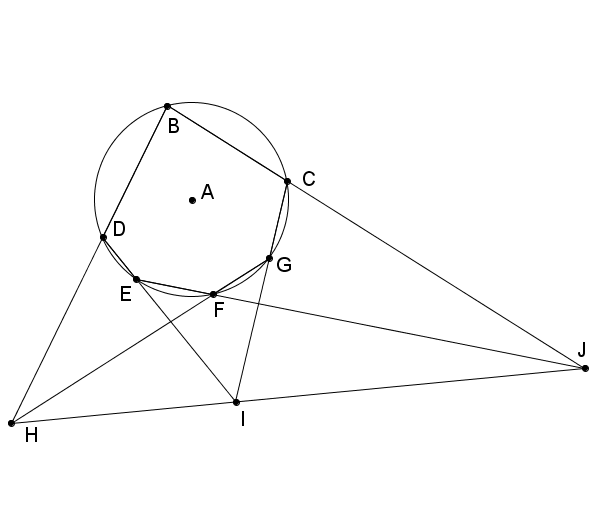}
    \end{minipage} & 1 & 0.249  & 8 & 8 \\
    \hline
    14 &
    \begin{minipage}[c]{0.25\textwidth}
            \includegraphics[width=2.8cm]{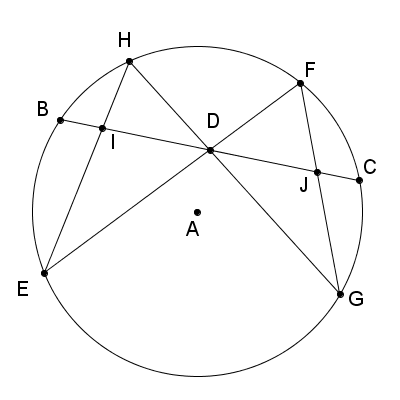}
    \end{minipage} & 0 & 0.171 & 6 &  4 \\
    \hline
    15 &
    \begin{minipage}[c]{0.25\textwidth}
        \includegraphics[width=2.8cm]{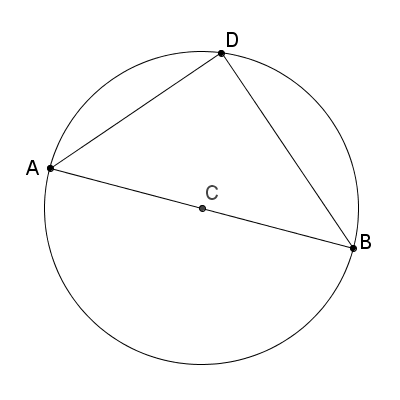}
    \end{minipage} & 0 & 0.2 & 2 & 2 \\
    \hline
    16 &
    \begin{minipage}[c]{0.25\textwidth}
            \includegraphics[width=2.8cm]{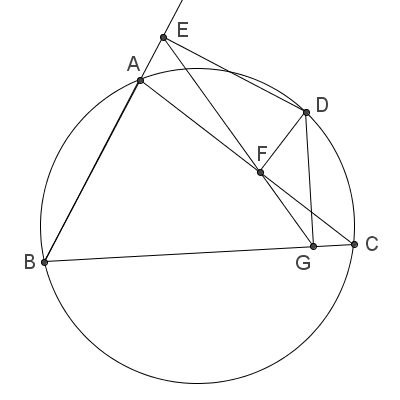}
    \end{minipage} & 0 & 0.171 & 2 & 2 \\
    \hline
    17 &
    \begin{minipage}[c]{0.25\textwidth}
            \includegraphics[width=2.8cm]{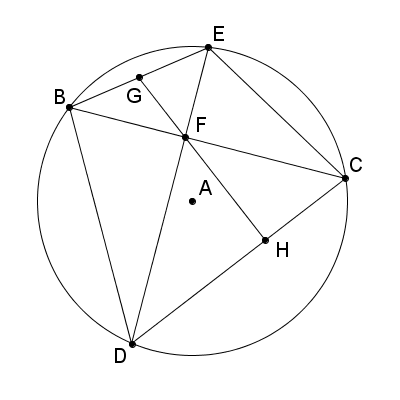}
    \end{minipage} & 0 & 0.187  & 4 & 4 \\
    \hline
    18 &
    \begin{minipage}[c]{0.25\textwidth}
        \includegraphics[width=2.8cm]{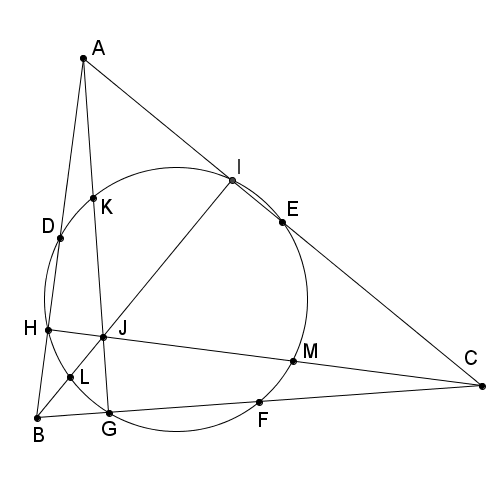}
    \end{minipage} & 1 & 0.281 & 7 & 7 \\
    \hline
    19 &
    \begin{minipage}[c]{0.25\textwidth}
        \includegraphics[width=2.8cm]{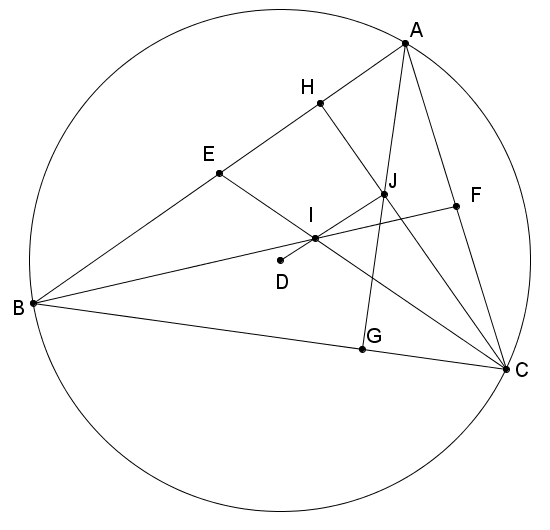}
    \end{minipage} & 10 & 0.312 & 3 & 3\\
    \hline
    20 &
    \begin{minipage}[c]{0.25\textwidth}
         \includegraphics[width=2.8cm]{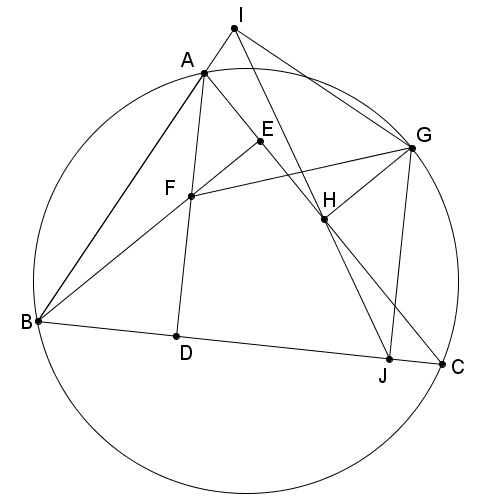}
    \end{minipage} & 4 & 0.312 & 7 & 6 \\
    \hline
    21 &
    \begin{minipage}[c]{0.25\textwidth}
         \includegraphics[width=2.8cm]{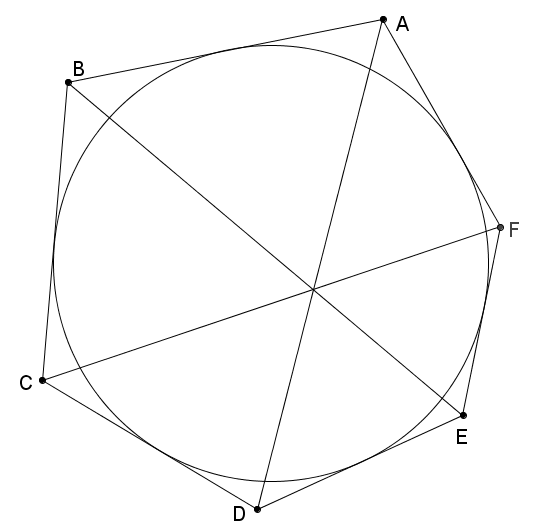}
    \end{minipage} & 11 & 0.451  & 10 & 0\\
    \hline
    22 &
    \begin{minipage}[c]{0.25\textwidth}
        \includegraphics[width=2.8cm]{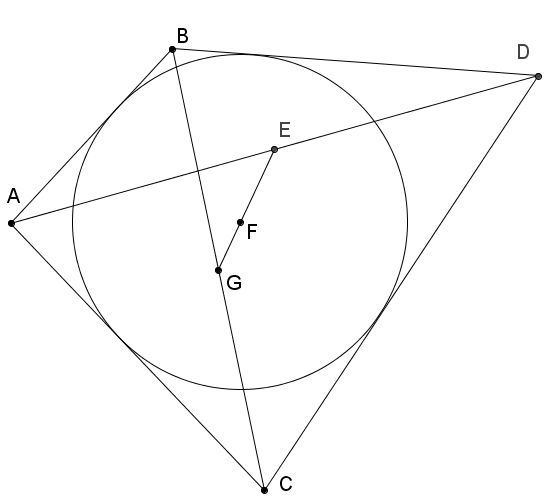}
    \end{minipage} & 5 & 0.219 & 8 & 0 \\
    \hline
    23 &
    \begin{minipage}[c]{0.25\textwidth}
         \includegraphics[width=2.8cm]{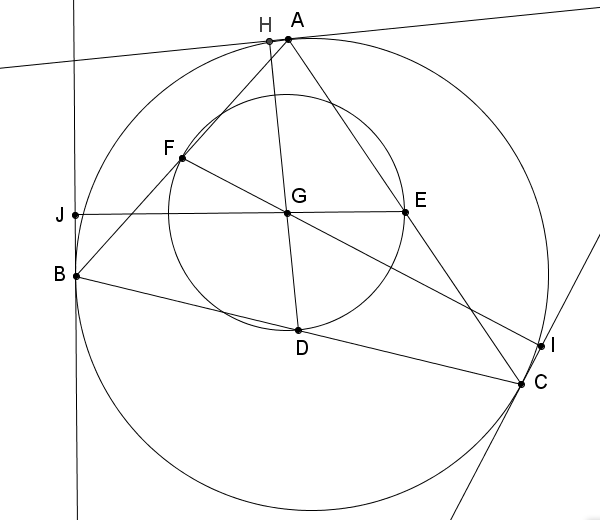}
    \end{minipage} & 27 & 0.453 & 9 & 0 \\
    \hline
\end{longtable}

In the test results, some undesired distance relations (such as
$\|AI\|=\|JM\|$ for the diagram image of the nine-point circle theorem with No.~18 and $\|FI\|=\|CH\|$ for that of Pappus' theorem with No.~1) are
retrieved due to insufficient accuracy of geometric object
recognition under large error tolerance. Generally speaking,
over-strict error tolerance may lead to the missing of useful
geometric relations for theorems that should be discovered, while
under-strict error tolerance may bring some spurious geometric
relations. Appropriate trade-off in the selection of error
tolerances for different images can help improve the completeness
and accuracy of geometric information retrieval.

For some images of diagrams (such as the image for
Th\'ebault's theorem with No.~12), the number
of generated candidate propositions is big because some branch
relations (e.g., $\|AC\|=\|CE\|, \|AC\|=\|AE\|\Rightarrow \|AE\|=\|CE\|$) are not ruled out.
For some other images of diagrams (such as the image for Morley's
theorem with No.~10 and that for Newton's theorem with No.~22), though
candidate propositions are generated successfully, the desired
theorems cannot be proved by using algebraic methods. This failure
of theorem proving is mainly for the following reasons.
 \begin{itemize}
 \item The automatically generated specifications of candidate propositions are not appropriate enough. For example, one of the generated candidate propositions for the image with No.~10 is Proposition(Morley$_1$, [$\|EF\| = \|DE\|$, $\angle ABE = \angle EBD$,  $\angle EBD = \angle DBC$,  $\angle FAE = \angle EAB$, $\angle CAF = \angle FAE$, $\angle BCD = \angle DCF$, $\angle DCF = \angle FCA$, $\|DF\|=\|EF\|$], [$\|DF\|=\|DE\|$]) in which only one relation is selected for the conclusion.
     The proposition should have been proved to be true because it is obvious that $\|EF\| = \|DE\|$ and $\|DF\|=\|EF\|$ imply $\|DF\|=\|DE\|$. However, symbolic computation with the algebraic relations expressing the hypothesis is so complicated that makes the program run out of memory. The candidate proposition fails to be a theorem because of inappropriate selection of relations for the hypothesis as well as the conclusion.
 \item The functions in GEOTHER we have used for automatic assignment
 of coordinates to points and ordering of variables are not well
 optimized.
 \end{itemize}
Thus the resulting algebraic expressions are much more complicated
than what could be produced with human optimization, so the involved
algebraic computations are made more complex as well.

\section{Related Work}\label{relatedwork}

\subsection{Geometric Information Retrieval}

Many methods have been proposed for shape recognition from images in
the last two decades. Some of them have improved the performance of
the traditional Hough transform by exploiting gradient
information~\cite{gradienthough} or using more effective voting
schemes~\cite{houghimp2}. Besides Hough transform, random algorithms
for the detection of lines and circles have been proposed
in~\cite{houghimp11,randomcircledetect}. Those algorithms save a
certain amount of storage space by first randomly computing a
candidate line or circle and then performing an evidence collecting
process to further determine whether the line or circle actually
exists. Note that most of the shape detection methods are used to
extract rough shapes of objects' edges from general images. Their
accuracy of recognition is not required to be very high. For our
purpose of recognizing geometric objects, it is crucial to use
OpenCV with numeric data (such as the coordinates of points) to
ensure that the accuracy of the detection results is sufficiently
high, so that geometric relations implied in images of diagrams can be
correctly determined through numeric computation.

\subsection{Geometric Theorem Discovery}

The reader may consult~\cite{survey,wu2,Recio09,gb2,wang95,wang96,wu} and references
therein for extensive studies on algebraic methods (based on
characteristic sets, triangular decomposition, and Gr\"{o}bner
bases) for automated proving and discovering of geometric theorems.
Here as examples we mention the open web-based tool~\cite{botana03}
developed for automatic discovery of theorems and relations in
elementary Euclidean geometry and the deductive database approach~\cite{deduction} proposed for searching all the properties implied
in given geometric configurations. In comparison with the existing
work, the capability of discovering nontrivial theorems or deductive
relations on geometric relations mined automatically from given
images of diagrams reflects the novelty of our approach.

\subsection{Other Related Work}

Besides coordinate-based algebraic methods, other methods for
automated theorem proving can also be incorporated into our approach
to verify the truth of candidate propositions. Such methods include
the area method, the full-angle method, the bracket algebra method,
methods based on Clifford algebra, axiom-based deductive methods,
and diagrammatic reasoning methods (see~\cite{diagram,survey,wang96}
and references therein). Some dynamic geometry software systems have
implemented specialized methods (e.g., randomized proving methods
in Cinderella~\cite{cinderella}) to prove theorems for constructed
diagrams, or interfaces with geometric theorem provers for
generating proofs diagrammatically~\cite{fleuriot,zheng} and exploring
knowledge in repositories of geometric constructions and
proofs~\cite{geothms}. A web-based library of problems in geometry
is being created for testing and evaluating methods and tools of
automated theorem proving~\cite{tgtp}. A new computational model for
computer assisted construction and reasoning of origami has been
well studied and used for proving some complicated
theorems~\cite{Ida}. Recently, proof assistants have been used
to interactively construct and verify proofs in geometry (see, e.g.,~\cite{coq}) and formal systems have established faithful models of proofs from
Euclid's \emph{Elements}, making use of diagrammatic reasoning (see, e.g.,~\cite{Avigad}).

\section{Conclusion and Future Work}\label{conclusion}

The approach proposed in this paper opens up a completely new route
for geometric knowledge discovery and reasoning: retrieve
characteristic information (geometric objects and their relations)
from simple and inexact data (images of diagrams), generate potential
knowledge (candidate propositions) from the retrieved information,
and discover profound knowledge (geometric theorems) and validate it by
means of automated reasoning (geometric theorem proving). The
success of our approach demonstrates the feasibility of
automatically acquiring formalized geometric knowledge in quantity
from a large scale of images of diagrams available in electronic
documents and resources and of efficiently managing such knowledge
in a retrievable structure with diagrams instead of ambiguous
statements in natural languages.

Our work is still ongoing. More experiments are being carried out
and more techniques and strategies are being developed to improve
the accuracy of retrieving geometric information from images of diagrams
and of ruling out branch relations and introducing derived relations, to
generate appropriate specifications of candidate propositions
heuristically, and to enhance the efficiency of geometric theorem
proving with optimal assignment of coordinates to points.

Currently, the images for experiments are produced from accurate diagrams drawn by using dynamic geometry software. We will extend our approach to deal with scanned and photographed images of hand-drawn diagrams in which the implied geometric relations are inexact. In this case, the
retrieval of geometric information becomes more difficult and requires more
specialized techniques. The
outcome of our study is expected to have practical applications in
those areas where geometric information retrieval, knowledge
discovery and management, and education are of concern.


\begin{thebibliography}{33}

\bibitem{Avigad} J. Avigad, E. Dean, and J. Mumma: A formal system for Euclid's \emph{Elements}. The Review of Symbolic Logic 2(4):700--768 (2009)
\bibitem{diagram} P. Balbiani and L. Fari\~nas del Cerro: Diagrammatic reasoning in projective
geometry. In: Logic, Language and Reasoning (H.J.\ Ohlbach and U.\ Reyle,
eds.), Trends in Logic 5, pp.\ 99--114, Kluwer, Dordrecht (1999)
\bibitem{botana03} F. Botana: A web-based intelligent system for geometric discovery. In:
Computational Science -- ICCS 2003, LNCS
2657, pp.\ 801--810, Springer, Berlin Heidelberg (2003)
\bibitem{houghimp11}
T.C. Chen and K.L. Chung: A new randomized algorithm for detecting
lines. Real-Time Imaging 7(6):473--481 (2001)
\bibitem{randomcircledetect}
T.C. Chen and K.L. Chung: An efficient randomized algorithm for
detecting circles. Computer Vision and Image Understanding 83(2):172--191 (2001)
\bibitem{survey} S.-C. Chou and X.-S. Gao: Automated reasoning in geometry, Handbook of Automated Reasoning, Volume\ I, Elsevier, North Holland (2001)
\bibitem{deduction} S.-C. Chou, X.-S. Gao, and J.-Z. Zhang: A deductive database approach to automated geometry theorem proving and discovering. Journal of Automated Reasoning 25(3): 219--246 (1996)
\bibitem{wu2} S.-C. Chou and D. Lin: Wu's method for automated geometry theorem proving and discovering. In: Mathematics mechanization and applications (X.-S. Gao and D. Wang, eds.), pp.\ 125--146. Academic Press, London (2000)
\bibitem{Recio09} G. Dalzotto and T. Recio: On protocols for the automated discovery of theorems in elementary geometry. Journal of Automated Reasoning 43(2):203--236 (2009)
\bibitem{hough}
R.O. Duda and P.E. Hart: Use of the Hough transformation to detect
lines and curves in pictures. Communications of Association for
Computing Machinery 15(1):11--15 (1972)
\bibitem{houghimp2}
 L.A.F. Fernandes and M.M. Oliveira: Real-time line detection through an improved Hough transform voting scheme. The Journal of the Pattern Recognition Society 41(1): 299--314 (2005)
\bibitem{gradienthough}
C. Galambos, J. Kittler, and J. Matas: Gradient based progressive
probabilistic Hough transform. Vision, Image and Signal Processing 148(3):158--165 (2001)
\bibitem{Ida} T. Ida, A. Kasem, F. Ghourabi, and H. Takahashi: Morley's theorem revisited: Origami construction and automated proof. Journal of Symbolic Computation 46(5):571--583 (2011)
\bibitem{cinderella} U. Kortenkamp: Foundations of dynamic geometry. Ph.D. thesis, pp.\ 60--72, ETH Z\"{u}rich (1999)
\bibitem{coq} N. Magaud, J. Narboux, and P. Schreck: Formalizing projective plane geometry in Coq. In: Automated Deduction in Geometry, LNAI 6301, pp.\ 141--162. Springer, Berlin Heidelberg (2011)
\bibitem{ppht} J. Matas, C. Galambos, and J. Kittler: Robust detection of lines using the progressive probabilistic Hough transform. Computer Vision and Image Understanding 78(1):119¨C-137 (2000)
\bibitem{gb2} A. Montes and T. Recio: Automatic discovery of geometry theorems using minimal canonical comprehensive Gr\"{o}bner systems. In: Automated Deduction in Geometry, LNAI 4869, pp.\ 113--138. Springer, Berlin Heidelberg (2007)
\bibitem{tgtp} P. Quaresma: Thousands of geometric problems for geometric theorem provers (TGTP). In: Automated Deduction in Geometry, LNAI 6877, pp.\ 169--181. Springer, Berlin Heidelberg (2011)
\bibitem{geothms} P. Quaresma and P. Jani\v ci\'c: GeoThms --- A web system for Euclidean constructive geometry. Electronic Notes in Theoretical Computer Science 174(2):35--48 (2007)
\bibitem{wang95} D. Wang: Elimination procedures for mechanical theorem proving
in geometry. Annals of Mathematics and Artificial Intelligence 13(1--2):1--24 (1995)
\bibitem{wang96} D. Wang: Geometry machines: from AI to SMC. In: Artificial Intelligence and Symbolic
Mathematical Computation (J. Calmet, J.A. Campbell, and J. Pfalzgraf, eds.), LNCS 1138, pp.\ 213--239. Springer, Berlin Heidelberg (1996)
\bibitem{wang} D. Wang: Elimination methods. Springer, Wien New York (2001)
\bibitem{fleuriot} S. Wilson and J.D. Fleuriot: Combining dynamic geometry, automated geometry theorem proving and diagrammatic proofs. In: Proceedings of the European Joint Conferences on Theory and Practice of Software (ETAPS), Satellite Workshop on User Interfaces for Theorem Provers (UITP), Edinburgh, UK (2005)
\bibitem{wu} W.-t. Wu: Mechanical theorem proving in geometries: Basic principles (translated from the Chinese by X. Jin and D. Wang). Springer, Wien New York (1994)
\bibitem{shi} K. Yano: The famous theorems of geometry (Chinese edition, translated by Y. Chen). Shanghai Scientific and Technical Publishers (1986)
\bibitem{zheng} Z. Ye, S.-C. Chou, and X.-S. Gao: Visually Dynamic Presentation of Proofs
in Plane Geometry. Journal of Automated Reasoning 45(3):213--241 (2010)
\bibitem{gbht} H.K. Yuen, J. Princen, J. Illingworth, and J.  Kittler: Comparative study of Hough transform methods for circle finding. Image and Vision Computing 8(1):71--77 (1990)
\bibitem{thinning} T.Y. Zhang and C.Y. Suen: A fast parallel algorithm for thinning digital patterns. Communications of the Association for Computing Machinery 27(3):236--239 (1984)


\bibitem{epsilon} Epsilon, \url{http://www-polsys.lip6.fr/~wang/epsilon/}. Accessed May 23 2014
\bibitem{smoothing}
Gaussian smoothing,
\url{http://en.wikipedia.org/wiki/Gaussian_blur}. Accessed May 23 2014
\bibitem{geogebra}
GeoGebra, \url{http://www.geogebra.org/cms/}. Accessed May 23 2014
\bibitem{geother} GEOTHER, \url{http://www-polsys.lip6.fr/~wang/GEOTHER/}. Accessed May 23 2014
\bibitem{dgs} List of interactive geometry software, \url{http://en.wikipedia.org/wiki/List_of_interactive_geometry_software}. Accessed May 23 2014
\bibitem{opencv}
OpenCV, \url{http://opencv.org/}. Accessed May 23 2014
\end{thebibliography}
\end{document}